\documentclass[journal]{IEEEtran}

\usepackage{algorithm}
\usepackage{algorithmic}
\newcommand{\etal}{\textit{et al.}}
\ifCLASSOPTIONcompsoc
  \usepackage[nocompress]{cite}
\else
  \usepackage{cite}
\fi
\ifCLASSINFOpdf
\else
\fi
\usepackage{times}
\usepackage{epsfig}
\usepackage{graphicx}
\usepackage{amsmath}
\usepackage{amssymb}
\usepackage{xcolor}
\usepackage{mathtools}
\usepackage{amsmath}
\usepackage{amsfonts}
\usepackage[normalem]{ulem}
\useunder{\uline}{\ul}{}
\usepackage{graphicx} 
\usepackage{tabularx} 
\usepackage{setspace}
\usepackage{pifont}
\usepackage{multirow}
\usepackage{subfig}
\usepackage{caption}

\newcommand{\changes}{\textcolor{black}}
\newcommand{\makesure}{\textcolor{black}}
\newcommand{\red}{\textcolor{black}}
\newcommand{\blue}{\textcolor{black}}
\newcommand{\green}{\textcolor{black}}

\newcommand{\revised}{\textcolor{black}}

\begin{document}
%
\title{\green{Stabilizing Adversarially Learned One-Class Novelty Detection Using Pseudo Anomalies}}
%
%
%
%

\author{Muhammad Zaigham Zaheer,
       Jin Ha Lee,
        Arif Mahmood,
        Marcella Astrid,
        and~Seung-Ik Lee
        
\IEEEcompsocitemizethanks{\IEEEcompsocthanksitem M. Zaigham Zaheer, Jin Ha Lee, Marcella Astrid and Seung-Ik Lee are with the University of Science and Technology and Electronics and Telecommunications Research Institute, Daejeon, Korea.
\protect\\
E-mail: see sites.google.com/view/cvml-ust/members 
\IEEEcompsocthanksitem Arif Mahmood is with the Information Technology University, Pakistan.}
\thanks{Manuscript received April 19, 2005; revised August 26, 2015.}}

%
%

\markboth{Journal of \LaTeX\ Class Files,~Vol.~14, No.~8, August~2015}%
{Shell \MakeLowercase{\textit{et al.}}: Bare Demo of IEEEtran.cls for Computer Society Journals}
%



\maketitle

\begin{abstract} 
\green{Recently, anomaly scores have been formulated using reconstruction loss of the adversarially learned generators and/or classification loss of discriminators. Unavailability of anomaly examples in the training data makes optimization of such networks challenging. Attributed to the adversarial training, performance of such models fluctuates drastically with each training step, making it difficult to halt the training at an optimal point. In the current study, we propose a robust anomaly detection framework that overcomes such instability by transforming the fundamental role of the discriminator from identifying real vs. fake data to distinguishing good vs. bad quality reconstructions.
For this purpose, we propose a method that utilizes the current state as well as an old state of the same generator to create good and bad quality reconstruction examples.
The discriminator is trained on these examples to detect the subtle distortions that are often present in the reconstructions of anomalous data.
In addition, we propose an efficient generic criterion to stop the training of our model, ensuring elevated performance.
Extensive experiments performed on six datasets across multiple domains including image and video based anomaly detection, medical diagnosis, and network security, have demonstrated excellent performance of our approach.}
\end{abstract}

\begin{IEEEkeywords}
Novelty detection, anomaly detection, adversarial learning, one-class classification, outliers detection, stabilizing adversarial models.
\end{IEEEkeywords}


%
\IEEEpeerreviewmaketitle

\section{Introduction}\label{sec:introduction}

%
%
%
%
\IEEEPARstart{A}{nomalies} are typically perceived as noise, erroneous observations or wrong measurements. However, \changes{in some cases,  learning about anomalies} may lead to a deeper understanding of the system being analyzed. For example, extremely few deflected alpha particles in the Rutherford gold foil experiment led to new understanding of the atomic structure of matter \cite{geiger1908scattering}. \changes{Similarly, in many applications,} deviations from the normal behavior are the actual objects of interest, detection of which can unleash the potential to address several challenging problems in pattern analysis and machine learning domains such as autonomous surveillance, medical diagnosis, datasets pre-processing, fraud detection, malicious attacks detection, and sports highlights generation.




Most of the anomaly detection systems attempt to learn the prototypical normal behavior of the data being observed. By modeling the gist of the dominant data trends, a reasoning is developed to evaluate unseen data and predict its `normalcy'. \blue{Therefore, anomaly detection is most commonly formulated as one-class classification (OCC) problem in which normal examples are used to train a novelty detection model} \cite{liu2018future_novelty,zhang2016video_novelty,luo2017revisit_novelty,hinami2017joint_novelty,sabokrou2017deep_novelty,hasan2016learning_novelty,smeureanu2017deep_novelty,ravanbakhsh2018plug_novelty,ravanbakhsh2017abnormal_novelty,wang2020novelty}. To this end, an encoder-decoder architecture such as denoising autoencoder is often used to learn the training class representations \cite{sabokrou2020deep,xu2015learning_denoise,xu2017detecting_denoise,vincent2008extracting_denoise}. Generally, in this scheme, training is carried out until the model starts producing good quality reconstructions of the normal class \cite{sabokrou2020deep, Shama_2019_ICCV_good}. \changes{During testing, the model is then expected to show high reconstruction loss corresponding to high anomaly scores for the samples outside the learned distribution.}  
However, relying only on the reconstruction capability of an autoencoder may not be a good strategy because, as the learning proceeds, it may also start well reconstructing the unseen data which can drastically degrade the anomaly detection performance \cite{Gong_2019_ICCV}. To overcome this, it becomes quite important to stop the training process appropriately before the autoencoder gets over trained in reconstructing the normal class. However, without anomaly examples, it can be cumbersome to define a good stopping criterion for these systems. 

\begin{figure}[t]
\begin{center}
\includegraphics[width=0.95\linewidth]{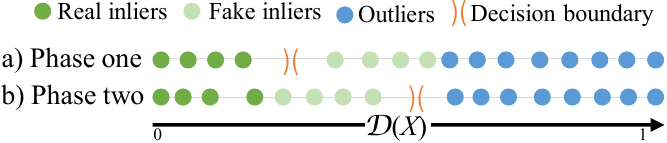}
\end{center}
\caption{\changes{a) An illustration of the behavior of a conventional discriminator, when used as it is for outlier detection problem. As it is trained to discriminate real from fake, the decision boundary lies in between the real inliers and fake inliers. b) The desideratum is to train a discriminator that can identify the reconstructions of inliers from the reconstructions of outliers.}}
\vspace{-2mm}
\label{fig:rebuttal_figure}
\end{figure}

With the recent development of Generative Adversarial Networks (GANs) \cite{goodfellow2014generative_gan}, some researchers explored the possibility of using adversarial training which can enhance the data regeneration quality  \cite{Shama_2019_ICCV_good,radford2015unsupervised_good_adversarial,pathak2016context_adversarial_good}. \changes{Once adversarial training is completed, the generator $\mathcal{G}$ is then decoupled from  the discriminator $\mathcal{D}$ to be used as a conventional reconstruction based anomaly detection model} \cite{ravanbakhsh2019training,ravanbakhsh2017abnormal_novelty}.
However, an improved regeneration capability of the generator may not guarantee improved anomaly detection performance because the \changes{ problem of autoencoders reconstructing the anomalous input has not been directly addressed. Another fundamental limitation of reconstruction based systems is that the reconstruction error is directly converted to anomaly score. In some systems, this score is computed and \textit{normalized} on a test video \cite{Gong_2019_ICCV,hasan2016learning_novelty}. The relatively higher values are then considered as corresponding to anomalies.}
\changes{This configuration, when applied to a test video without anomalies, will still be biased towards labeling some elements as anomalous, thus reducing the overall efficacy of the system.}

\begin{figure}[t]
\begin{center}
   \includegraphics[width=0.75\linewidth]{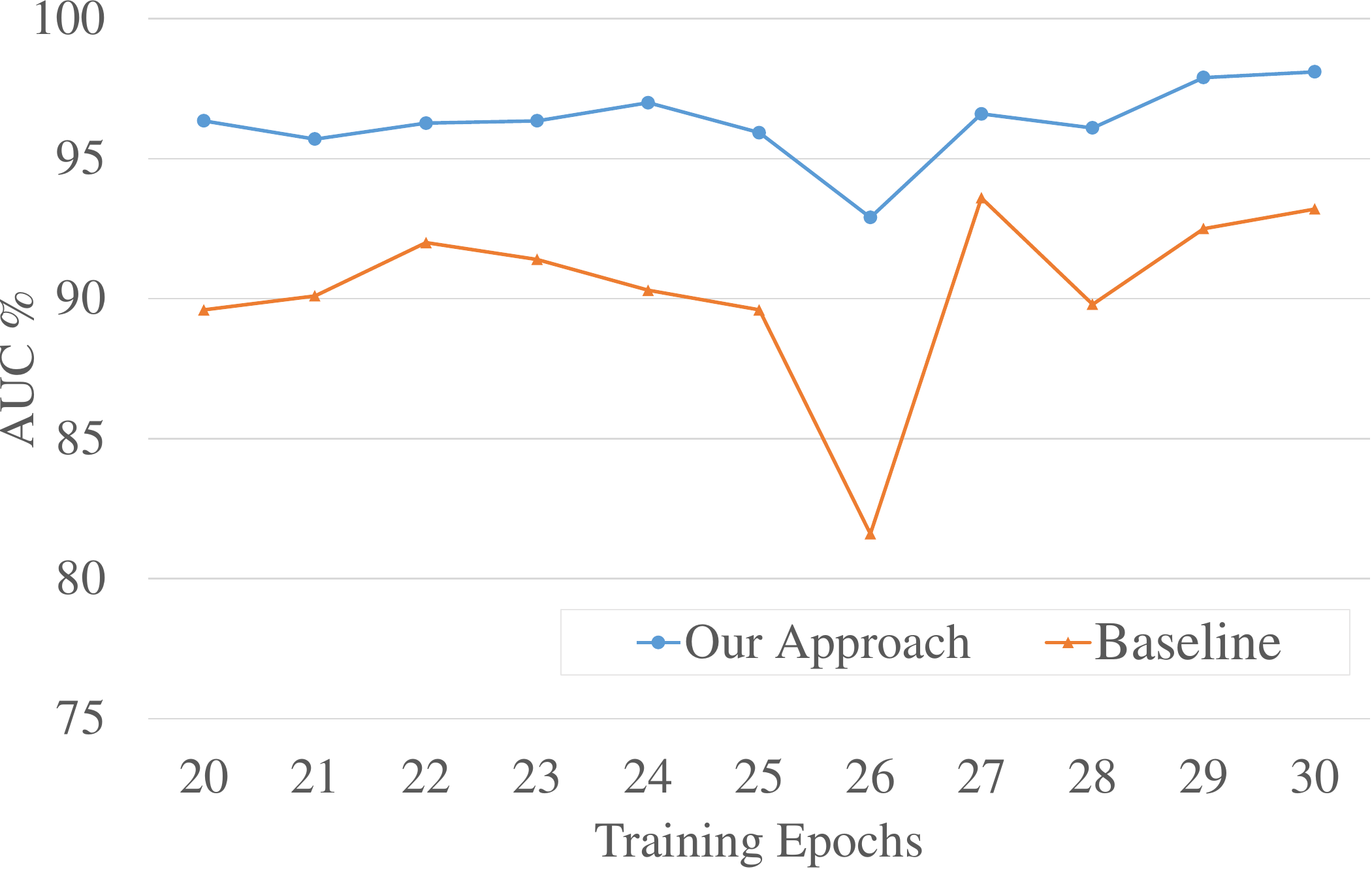}
\end{center}
   \caption{Dynamics of AUC performance over training epochs: The baseline ($\mathcal{G}$+$\mathcal{D}$) shows fluctuations while our approach not only shows stability across various epochs but also yields higher AUC.}
    \vspace{-2mm}
\label{fig:AUC_Comparison}
\end{figure}

\blue{A natural drift in the domain of adversarial learning for anomaly detection is the idea of utilizing discriminator $\mathcal{D}$ together with the generator $\mathcal{G}$ for anomaly scoring. The intuition is to reap maximum benefits of the adversarial training by involving both $\mathcal{G}$ and $\mathcal{D}$ instead of only $\mathcal{G}$.} \changes{Sabokrou \etal \cite{sabokrou2020deep} \:proposed to input a reconstructed image to $\mathcal{D}$ and label it anomalous if classified as fake.} This configuration makes the anomaly scoring simple as $\mathcal{D}$ is already trained to regress scores within a certain range, mitigating the need of further formulating or normalizing the anomaly scores. Thus, in contrast to the previous systems, such models can predict scores individually for each test element. However, this configuration also brings along new challenges commonly associated with adversarial architectures. 
First, \changes{as illustrated in Fig. \ref{fig:rebuttal_figure}a, the conventional way of training $\mathcal{D}$ to discriminate between real and fake compels it to draw a boundary within inlier region where real and fake inliers meet. However, this is not desirable in outlier detection since  the boundary should be in between the inliers and the outliers, as chalked out in Fig. \ref{fig:rebuttal_figure}b.}
\blue{Secondly, the performance of such adversarial architectures is strictly dependent on the criteria of when to stop the training \cite{sabokrou2020deep}.
In the case of halting prematurely, $\mathcal{G}$ may remain under-trained whereas in the case of over-training, $\mathcal{D}$ may not perform well because of the real-looking fake data.} Furthermore, since $\mathcal{G}$ and $\mathcal{D}$ compete against each other, both may not  converge at the same time.
We experimentally observed that the performance of such a system having $\mathcal{G}$+$\mathcal{D}$ trained as a collective model for anomaly detection fluctuates significantly  (Fig. \ref{fig:AUC_Comparison}), making it difficult to formulate a good stopping point without the availability of outlier examples.

To mitigate these issues, we propose \blue{to transform the conventional role of $\mathcal{D}$ from identifying real and fake to detecting good and bad quality reconstructions. For anomaly detection, this is a highly desirable attribute of $\mathcal{D}$ because a trained $\mathcal{G}$ may leave more distortions in the reconstructions of abnormal data compared to normal data. To this end, we formulate a two phase training process. Phase one is similar to the conventional approach of training an adversarial denoising autoencoder \cite{pathak2016context_adversarial_good,xu2017detecting_denoise,vincent2008extracting_denoise}.} However, during early training iterations \changes{of phase one}, an initial version of $\mathcal{G}$ is preserved as $\mathcal{G}_{o}$ to be later used, while the training  of $\mathcal{G}$+$\mathcal{D}$ model continues. Once $\mathcal{G}$ achieves low reconstruction loss, we save it as $\mathcal{G}_{n}$ and proceed to phase two. \blue{In this phase, only $\mathcal{D}$ is trained on several good and bad quality reconstruction examples. Good quality examples are taken from the normal training data as well as the data reconstructed by $\mathcal{G}_{n}$. Bad quality examples are generated by employing $\mathcal{G}_{o}$ as well as by using a newly proposed pseudo anomaly module which utilizes the normal training data to produce \textit{anomaly-like} samples. The  aim of phase two becomes to evolve $\mathcal{D}$ in such a way that it can \changes{adopt the behavior illustrated in Fig. \ref{fig:rebuttal_figure} and} accurately identify the reconstructions coming from anomalous and normal data at test time. 
Fig. \ref{fig:AUC_Comparison} shows frame-level area under the curve (AUC) performance comparison of the proposed framework and the conventional $\mathcal{G}+\mathcal{D}$ model over several epochs of training on UCSD Ped2 dataset \cite{chan2008ucsd}.
Although the baseline provides reasonable AUC occasionally, its overall performance  fluctuates substantially even between two arbitrary consecutive epochs. In contrast, our proposed approach not only demonstrates noticeably superior performance but also yields stability across several training epochs.}

A preliminary version of this work was recently presented in the Conference on Computer Vision and Pattern Recognition (CVPR) 2020 \cite{zaheer2020old}. This current work is a substantial extension of the conference version. \changes{First,} we extend the idea of pseudo anomaly module by introducing three different fusion methods: early, late and latent fusion. \changes{Secondly, rather than handpicking a stopping criterion, we propose a reconstruction loss based generic formulation which further enhances the robustness of our method. Thirdly, we broaden the evaluation of our approach and report detailed results and analysis on
MNIST \cite{mnist}, Caltech-256 \cite{griffin2007caltech}, Ped2\cite{chan2008ucsd}, Wisconsin Breast Cancer \cite{Wolberg1992wbcd}, Inbreast \cite{moreira2012inbreast}, and KDDCCUP \cite{tavallaee2009detailed} datasets across four different domains i.e. } image outliers detection, video anomaly detection, medical diagnosis, and network security. 
In summary, the contributions of our work are as follows:

\begin{enumerate}
  \item  We extensively  analyze the impacts of using $\mathcal{G}+\mathcal{D}$ formulation for anomaly detection  and report the instability of such systems.
   \item {We change the role of $\mathcal{D}$ to an anomaly detector by creating good and bad quality reconstruction examples using two versions, $\mathcal{G}_{n}$ and $\mathcal{G}_{o}$, of the generator. This not only mitigates the instability caused by adversarial training but also elevates the overall performance.}
  \item We analyze several pseudo anomaly modules to enhance the training of $\mathcal{D}$ \changes{as anomaly detector}.
  \item We devise a stopping criterion for our approach which enables it to select efficient $\mathcal{G}_{o}$ and $\mathcal{G}_{n}$ models. 
 \item We provide extensive performance evaluation of our approach on \changes{several datasets across four different domains} including image outlier detection, video anomaly detection, medical diagnosis and network security.
  \end{enumerate}




\section{Related Work}
\blue{Anomaly detection is usually formulated as novelty detection problem \cite{liu2018future_novelty,zhang2016video_novelty,luo2017revisit_novelty,hinami2017joint_novelty,xia2015learning_novelty_fig5,sultani2018real_novelty,sabokrou2017deep_novelty,bergadano2019keyed,hasan2016learning_novelty,smeureanu2017deep_novelty,ravanbakhsh2018plug_novelty,pang2021deep,pang2018learning} in which the known normal class is learned by a model. At test time, the model then identifies outliers as anomalous. To tackle the problem, some researchers proposed the utilization of object trackers to identify anomalies
\cite{basharat2008learning_realworld7,medioni2001event_twostream34,piciarelli2008trajectory_twostream36,zhang2009learning_twostream53}.
However, using such handpicked features may degrade the performance of a novelty detector significantly. With the ever increasing popularity of deep learning, Smeureanu \etal \cite{smeureanu2017deep_novelty} proposed the idea of utilizing deep features extracted using pre-trained convolution networks to train the conventional one-class classifiers. However, if trained on unrelated datasets, a feature extractor may limit the overall performance of the system.}

\blue{Data reconstruction based methods are a relatively new addition to the field of anomaly detection \cite{Gong_2019_ICCV,ren2015unsupervised,xu2015learning_denoise,ionescu2019object,Nguyen_2019_ICCV,nguyen2019hybrid,xu2017detecting_denoise,sabokrou2017deep_novelty,zhou2017anomaly}. These approaches utilize an unsupervised generative network to learn normal data representations. By assuming that a generator is unable to reconstruct outliers, Ravanbakhsh \etal\cite{ravanbakhsh2018plug_novelty} proposed to use the loss of a reconstructor for detecting anomalous events. Ionescu \etal \cite{ionescu2019object} proposed an approach combining object detectors and convolutional autoencoders to encode normal appearance and motion representations. Xu \etal \cite{xu2015learning_denoise,xu2017detecting_denoise} formulated a one-class SVM based approach trained using deep features from stacked autoencoders. The authors in \cite{Nguyen_2019_ICCV,nguyen2019hybrid} trained a cascaded decoder to learn appearance as well as motion from normal videos. However, in all these schemes, only a generator is employed for training as well as detection. }

\blue{\changes{Alternatively,} Pathak \etal \cite{pathak2016context_adversarial_good} proposed the utilization of a discriminator during training to enhance the quality of reconstruction. Nevertheless, at test time, the discriminator is decoupled and only generator is employed to compute anomaly scores. On the other hand, usage of a unified generator and discriminator model for anomaly scoring is proposed by Sabokrou \etal\cite{sabokrou2020deep}. Although it demonstrates promising results, the model is often unstable and the performance relies heavily on the stopping criteria of the training. Shama \etal\cite{Shama_2019_ICCV_good} recently proposed the idea of adversarial feedback loop in which output of the discriminator is used to enhance the quality of generated images. Although not related to anomaly detection, the approach provides a compelling intuition to utilize both components of an adversarial network for an elevated performance.}

\blue{Although utilizing an adversarial autoencoder network as baseline, our work is substantially different from the methods in\cite{Gong_2019_ICCV,ionescu2019object,Nguyen_2019_ICCV,nguyen2019hybrid,xu2017detecting_denoise,sabokrou2020deep} as we explore the possibility of employing a unified generator and discriminator model for anomaly scoring at test time. The most related approach to ours is by Sabokrou \etal\cite{sabokrou2020deep} as they also explore the unconventional usage of discriminator for scoring. 
\changes{Specifically, they train a conventional adversarial network and use the trained discriminator as it is for anomaly detection. In contrast, we propose the utilization of an old state of the generator} to transform the ultimate role of the discriminator from detecting fake and real to identifying good and bad quality reconstructions. This way, although initially trained adversarially, our framework ultimately aligns the discriminator to complement generator towards anomaly detection.}

Our method may also be seen as a variant of negative learning \red{\cite{munawar2017limiting}} in which few anomaly examples along with the inlier data are provided to the system during training to enhance the decision boundary. We also acknowledge a recent introduction of several weakly supervised methods which use both normal and anomalous examples for training \cite{zaheerclaws,sultani2018real_novelty,zaheer2020cleaning,zaheer2020self,zhong2019graph}. Mostly popular for videos, these methods utilize video-level weak supervision to train the frame-level prediction models. In contrast, we do not utilize any anomaly examples from the dataset for training. Rather, we stick to the conventional protocol of one-class classification and propose to generate pseudo anomaly examples from the normal training data.
%

\section{Proposed Adversarially Learned Novelty Detection Framework} \label{section:method}
In this section, we present our proposed novelty detection framework for one-class classification.
Although most adversarial training based approaches discard discriminator after training, we attempt to make use of it for producing anomaly scores. However, as shown in Fig. \ref{fig:AUC_Comparison}, we observe that the performance of a naive $\mathcal{G}+\mathcal{D}$ model (referred as baseline) is unstable due to adversarial nature of the training. 
In order to develop a stable and more robust novelty detection system, we propose a two-phase training approach. 
During phase one, a more evolved version of the generator generating good quality reconstructions and a less evolved version generating bad quality reconstructions are selected. 
During phase two, we utilize these generators to redefine the role of the discriminator from discriminating between real and fake examples to distinguishing between good and bad quality reconstructions. Additionally, to make the framework more robust, we also propose a pseudo anomaly generation module. 

In contrast to existing approaches, the performance of our system is not prone to over-training or under-training of the generator/discriminator. It is because, in phase two we stop adversarial training and force the discriminator to become consistent with the generator towards the goal of anomaly detection. This helps the overall system in producing stable results across several training epochs and achieving superior results. Our proposed framework is generic and can be integrated with many existing one-class novelty detection systems across several different domains.  


\begin{figure*}
\begin{center}
  \includegraphics[width=1\linewidth]{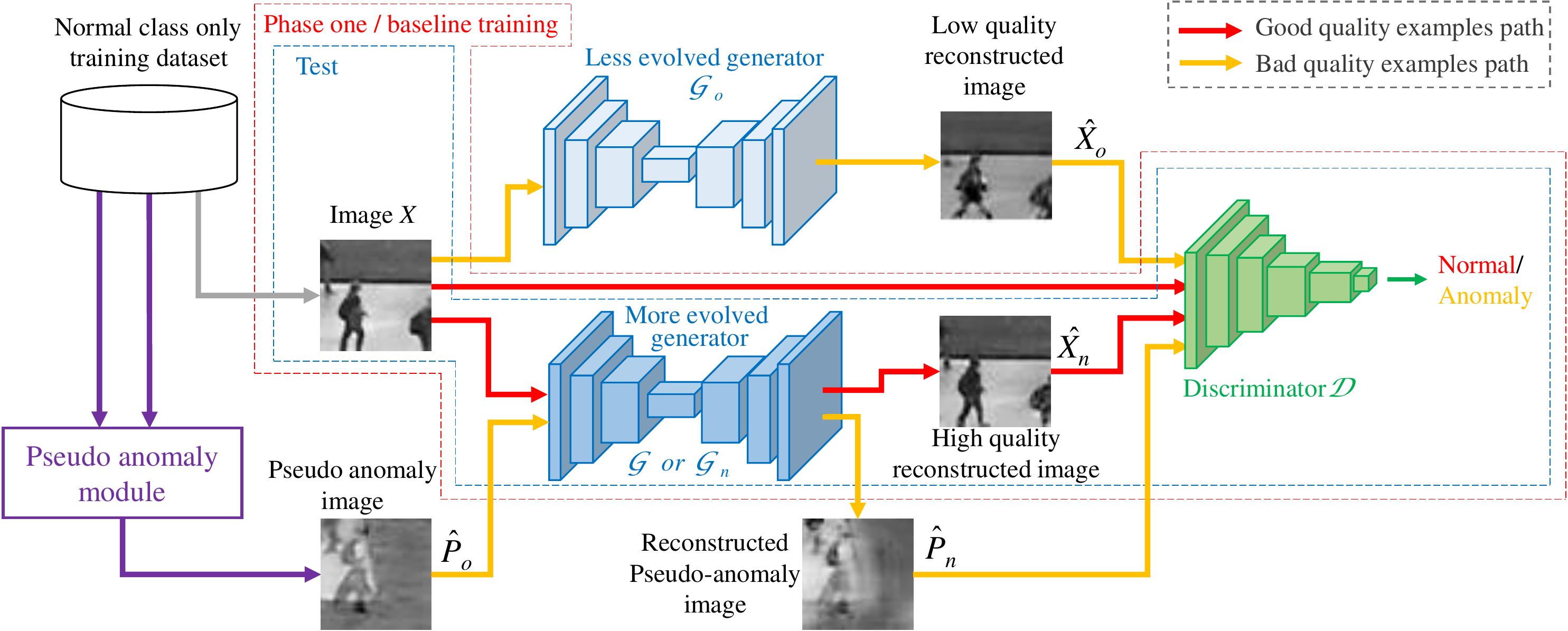}
   \caption{\blue{Our proposed adversarially learned novelty detection framework. Phase one is the baseline training, carried out to achieve a reasonably evolved state of $\mathcal{G}$ and $\mathcal{D}$. Two states of $\mathcal{G}$, a less evolved $\mathcal{G}_{o}$ and a more evolved $\mathcal{G}_{n}$, are frozen during this phase. Then, in phase two, only $\mathcal{D}$ is trained to distinguish between good and bad quality reconstruction examples. Good quality examples correspond to real training images as well as the training images reconstructed through $\mathcal{G}$. Whereas, bad quality examples are created using $\mathcal{G}_{o}$ as well as the newly introduced pseudo anomaly module. This module assists $\mathcal{D}$ in learning the underlying patterns of the anomalous input reconstructions. At test time, only $\mathcal{G}_n$ and $\mathcal{D}$ are used for inference and the output of $\mathcal{D}$ is directly considered as anomaly score.}}
\label{fig:architecture}
\end{center}
\vspace{-4mm}
\end{figure*}

\subsection{Architecture}
\blue{Our base system, consisting of a generator $\mathcal{G}$ which is a typical denoising autoencoder \cite{sabokrou2020deep} and a discriminator $\mathcal{D}$, learns one-class data representations in an unsupervised adversarial fashion. The model plays a mini-max game to optimize the following objective function:}
\begin{equation}
\begin{multlined}
\underset{\mathcal{G}}{\text{min}} \: \underset{\mathcal{D}}{\text{max}} \:  \Bigl(\mathbb{E}_{X \sim p_t}[\text{log}(1-\mathcal{D}(X))] \\ + \mathbb{E}_{\tilde{X} \sim p_t + \mathcal{N}_\sigma}[\text{log}(\mathcal{D}(\mathcal{G}(\tilde{X})))]\Bigr),
\end{multlined}
\label{eq:jointGAN}
\end{equation}
\changes {where $X$ is a sample drawn from a real data distribution $p_t$ and $\tilde{X}$ is the sample $X$ with added noise $\epsilon$, which is sampled from a normal distribution $\mathcal{N}_\sigma$ with  standard deviation $\sigma$ such that $\mathcal{N}_\sigma$:}  $\tilde{X} = (X \sim p_t) + (\epsilon \sim \mathcal{N}(0, \sigma^2I))$.  The generator $\mathcal{G}$ tries to minimize this objective function by removing the noise $\epsilon$, while the discriminator $\mathcal{D}$ tries to maximize it by predicting 0 for the real data $X$ and 1 for the regenerated data $\mathcal{G}(\tilde{X})$. 
Our model extends this baseline by redefining the role of the discriminator from identifying real data and noisy data reconstructions to discriminating between good and bad quality examples.

The good quality examples are taken from the real data distribution $p_t$ as well as the reconstructions $\hat{X}_n$ of the real data $X$ by a more evolved generator $\mathcal{G}_n$ (shown as solid red paths in Fig. \ref{fig:architecture}). The bad quality examples are obtained by either using a less evolved generator $\mathcal{G}_{o}$ or a pseudo anomaly module (shown as solid yellow paths). $\mathcal{G}_{o}$ generates distorted examples which are perturbations of the real data distribution $p_t$. In the perspective of $\mathcal{G}_n$, the output of $\mathcal{G}_{o}$  may be considered as reconstructed anomalous samples. Thus without explicitly using any out-of-class data, the behavior of $\mathcal{G}_n$ on the outlier data is indirectly mimicked to train the discriminator. We also propose a pseudo anomaly module that assists $\mathcal{D}$ in directly learning the behavior of $\mathcal{G}_n$ in the case of an unusual or anomalous input. To this end, we create pseudo anomaly examples by fusing randomly selected samples from the real data distribution $p_t$ and distorting them using $\mathcal{G}_o$. These distorted examples are then regenerated using $\mathcal{G}_n$ to be used as reconstructed anomaly examples in training $\mathcal{D}$. This explicitly helps $\mathcal{D}$ in learning the underlying patterns of  reconstructed  anomalous  data,  which  results  in  a more robust anomaly detection model. 

Thus, by redefining the learning paradigm of $\mathcal{D}$, we enable it to get aligned with the conventional philosophy of the generative one-class learning models in which the reconstruction quality of the data from a known class is better than the data from an unknown or anomalous class. 



\begin{figure*}[t]
\begin{center}
   \includegraphics[width=1\linewidth]{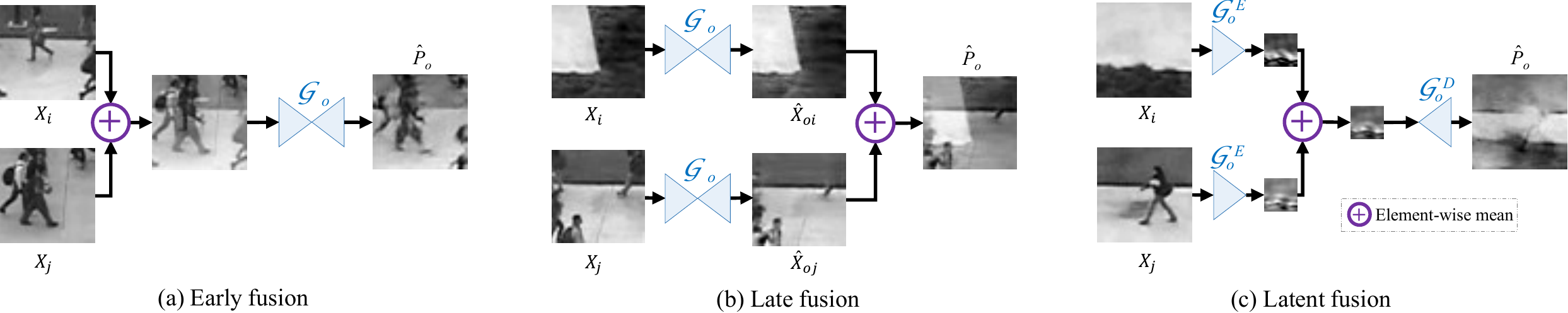}
\end{center}
   \caption{The three variants of our proposed pseudo anomaly module for generating pseudo anomalies $\hat{P}_{o}$. (a) in Early fusion, two input data instances are averaged and passed through $\mathcal{G}_{o}$ to obtain $\hat{P}_{o}$. (b) in Late fusion, two data instances are reconstructed separately using $\mathcal{G}_{o}$ and the outputs are averaged to form $\hat{P}_{o}$. (c) in Latent fusion, latent embeddings of the two input instances are computed using the encoder $\mathcal{G}_{o}^E$ of $\mathcal{G}_{o}$ and averaged. The resultant embedding is then passed through the decoder $\mathcal{G}^D_{o}$ of $\mathcal{G}_{o}$ to obtain $\hat{P}_{o}$. 
   }
\label{fig:pseudo_reconstruction_module}
\end{figure*}

\subsection{Training Phase One }
\blue{Our model (Fig. \ref{fig:architecture}) is trained in two phases. 
Phase one is identical to the previous approaches training an adversarial one-class classifier based on denoising autoencoder
 \cite{sabokrou2020deep}. In this phase, $\mathcal{G}$ reconstructs the real data $X$ to generate $\hat{X}_o$.
 Then $\mathcal{D}$ is trained to identify real and fake data, failure or success of which then becomes a supervision signal for $\mathcal{G}$. As burn-in stage, this unsupervised training is carried out for a few epochs.} Overall, phase one minimizes the following loss function:
\begin{equation}
\mathcal{L} = \mathcal{L}_{\mathcal{G}+\mathcal{D}} + \lambda\mathcal{L}_R,
\label{eq:lambda}
\end{equation}
where $\mathcal{L}_{\mathcal{G}+\mathcal{D}}$ is the loss function of our joint training objective defined in Eq. \eqref{eq:jointGAN} and
\begin{equation} \label{eq:lr}\mathcal{L}_R = || X - \mathcal{G}(\tilde{X}) ||^2\end{equation} is the reconstruction loss. 
Moreover, $\lambda$ is a loss balancing variable.

Quasi ground truth for the discriminator in phase one training is defined as:
\begin{equation}
    GT_{phase\_one}= 
\begin{cases}
    0      & \text{if input is $X$,}\\
    1      & \text{if input is $\mathcal{G}(\tilde{X})$}.
\end{cases}
\end{equation}

\noindent\textbf{Selection of $\mathcal{G}_o$ and $\mathcal{G}_n$ :}
Due to the adversarial setting, reconstruction loss of $\mathcal{G}$ fluctuates in phase one.
We save a less evolved low reconstruction quality generator model $\mathcal{G}_{o}$ and a more evolved generator model $\mathcal{G}_{n}$ to be used later in phase two of the training. For the selection of $\mathcal{G}_o$, a particular low epoch model is not essential. Rather a version of $\mathcal{G}$ with relatively higher reconstruction loss, $\mathcal{L}_R$ (consequently lower reconstruction quality) can be used as $\mathcal{G}_o$. More specifically, as shown in Algorithm 1, we aim $\mathcal{G}_{o}$ to be better than a randomly initialized version of $\mathcal{G}$ while having a higher $\mathcal{L}_R$ than $\mathcal{G}_n$. 
For a given $\mathcal{G}_{o}$, the corresponding $\mathcal{G}_{n}$ may be selected based on the following criterion:
\begin{equation}
    \frac{\mathcal{L}_R (\mathcal{G}_o)}{\mathcal{L}_R (\mathcal{G}_n)} > \eta,
    \label{eq:eta}
\end{equation}
where $\eta \ge 1$ is a threshold which is  used for the selection of $\mathcal{G}_n$ in a generic setting (see Section \ref{sec:discussion} for more details).


\begin{algorithm}
\caption{Selection of $\mathcal{G}_{o}$ and $\mathcal{G}_{n}$ in Phase One}
Initialize $\mathcal{G}$ \& $\mathcal{D}$ Randomly
\begin{algorithmic}
\REQUIRE iterate = True
\STATE Train $\mathcal{G} + \mathcal{D}$ for $p$ epochs \changes{(burn-in stage)}
\STATE $\mathcal{G}_{o} \leftarrow \mathcal{G}$
\STATE $\mathcal{G}_{n} \leftarrow \mathcal{G}$
\WHILE{(iterate = True)}
\STATE Train $\mathcal{G} + \mathcal{D}$ for one iteration
\IF{$\mathcal{L}_R (\mathcal{G}_{o}) < \mathcal{L}_R (\mathcal{G})$}
\STATE $\mathcal{G}_{o} \leftarrow \mathcal{G}$
\ELSE[$\mathcal{L}_R (\mathcal{G}_{n}) > \mathcal{L}_R (\mathcal{G})$]
\STATE $\mathcal{G}_{n} \leftarrow \mathcal{G}$
\ENDIF
\IF{$\mathcal{L}_R (\mathcal{G}_{o}) / \mathcal{L}_R (\mathcal{G}_{n}) > \eta$}
\STATE iterate $\leftarrow$ False
\ENDIF
\ENDWHILE
\STATE Begin phase two training using $\mathcal{G}_{o} \& \mathcal{G}_{n}$
\end{algorithmic}
\end{algorithm}

\subsection{Training Phase Two}
\label{subsection:phasetwo}
\changes{In phase two, we use the generator models $\mathcal{G}_o$ and $\mathcal{G}_n$ obtained in phase one to update the discriminator $\mathcal{D}$. In this phase, the role of $\mathcal{D}$ is} changed from discriminating between real and fake data to  differentiating between good and bad quality reconstructions. Such transformation of $\mathcal{D}$ makes it more suitable for one-class classification problems such as anomaly detection. 

The essence of phase two training is to provide examples of good and bad quality reconstructions to $\mathcal{D}$ with the goal of making it learn about the output that $\mathcal{G}_n$ may produce in the case of an unusual input. 
Details of phase two training are discussed next:

\subsubsection{Good and Bad Quality Examples} 
As good quality examples, $\mathcal{D}$ is provided with real data $X$ and the high quality reconstructed data: $\hat{X}_n = \mathcal{G}_{n}(X)$. \changes{Note that we do not add noise to X in this phase as the $\mathcal{G}_{n}$ is used only for inference.}

As bad quality examples, $\mathcal{D}$ is provided with low quality reconstructions $\hat{X}_{o}=\mathcal{G}_{o}(X)$. 
In addition, we also generate bad quality reconstruction examples by using a pseudo anomaly module, which is formulated with different combinations of $\mathcal{G}_{o}$ to create pseudo-anomalies $\hat{P}_o$.

\subsubsection{Pseudo Anomaly Module} 
In this work, we consider three different approaches to obtain pseudo anomalies as shown in Fig.~\ref{fig:pseudo_reconstruction_module} and described below.

\noindent\textbf{Pseudo anomaly generation by early fusion:}
The randomly selected samples $X_i$ and $X_j$ are fused in image space using element-wise mean and the resultant is reconstructed using $\mathcal{G}_{o}$.

\begin{equation}
\hat{P}_o = \mathcal{G}_{o}\left(\frac{X_i+X_j}{2}\right), ~~~ 
i \neq j
\label{eq:early_fusion}
\end{equation}

\noindent\textbf{Pseudo anomaly generation by late fusion:} Two distinct samples $X_i$ and $X_j$ are randomly selected from the training dataset. A pseudo anomaly $\hat{P}_o$ is then generated as the element-wise average of low quality reconstructions:

\begin{equation}\label{eq:fake_anomaly}
\hat{P}_o = \frac{\mathcal{G}_{o}(X_i) + \mathcal{G}_{o}(X_j)}{2}  = \frac{\hat{X}_{oi} + \hat{X}_{oj}}{2},  ~~~ 
i \neq j
\end{equation}

\noindent\textbf{Pseudo anomaly generation by latent space fusion:}
Two randomly selected samples $X_i$ and $X_j$ are input to $\mathcal{G}_{o}^E$, the encoder of $\mathcal{G}_{o}$, and latent representations of both samples are computed. Fusion is performed in the latent space by taking element-wise mean and the resultant representation is then decoded using $\mathcal{G}_{o}^D$, the decoder of $\mathcal{G}_{o}$.
\begin{equation}
\hat{P}_o = \mathcal{G}_o^D\left( \frac{\mathcal{G}_o^{E}(X_i) + \mathcal{G}_o^{E}(X_j)}{2} \right),  ~~~ 
i \neq j
\label{eq:latent_fusion}
\end{equation}

The output obtained using either of the pseudo anomaly modules may contain diverse variations which are completely unknown to both $\mathcal{G}_n$ and $\mathcal{D}$ models.
Finally, as the last step in this pipeline, in order to mimic the behavior of $\mathcal{G}_n$ on unusual input data, pseudo anomaly examples $\hat{P}_o$ are reconstructed using $\mathcal{G}_n$ to obtain $\hat{P}_n$:
\begin{equation}\label{eq:pseudo_reconstruction}
\hat{P}_{n} = \mathcal{G}_n(\hat{P}_o).
\end{equation}
Example images at several intermediate steps of our framework can be seen in Fig.~\ref{fig:pseudo_reconstruction_module} and~\ref{fig:example_images}. 

\subsubsection{Training the Discriminator} Objective function of the model in phase two takes the form: 
\begin{equation}
\begin{multlined}
\underset{\mathcal{D}}{\text{max}}\:\Bigl(\alpha\mathbb{E}_{X}[\text{log}(1-\mathcal{D}(X))] +\\(1-\alpha)\mathbb{E}_{\hat{X}_n}[\text{log}(1-\mathcal{D}(\hat{X}_n))] + \beta\mathbb{E}_{\hat{X}_{o}}[\text{log}(\mathcal{D}(\hat{X}_{o}))] +\\(1-\beta)\mathbb{E}_{\hat{P}_o}[\text{log}(\mathcal{D}(\hat{P}_n))]\Bigr),
\end{multlined}
\label{eq:alpha}
\end{equation}
where $\alpha$ and $\beta$ are the weighing parameters. For the phase two training, ground truth takes the form: 
\begin{equation}
    GT_{phase\_two}= 
\begin{cases}
    0      & \text{if input is $X$ or $\hat{X}_n$,}\\
    1      & \text{if input is $\hat{X}_o$ or $\hat{P}_n$}.
\end{cases}
\end{equation}
\subsection{Testing}
At test time, as shown in Fig.~\ref{fig:architecture}, only $\mathcal{G}_{n}$ and $\mathcal{D}$ are utilized for one-class classification (OCC). Final classification decision for an input data $X$ is given as:
\begin{equation}
OCC = 
\begin{cases}
    \text{normal class}      & \text{if $\mathcal{D}(\mathcal{G}_{n}(X)) < \tau$,}\\
    \text{anomaly class}      & \text{otherwise,}
\end{cases}
\label{eq:threshold}
\end{equation}
where $\tau$ is a predefined threshold. 

\begin{figure}[t]
\begin{center}
\includegraphics[width=1\linewidth]{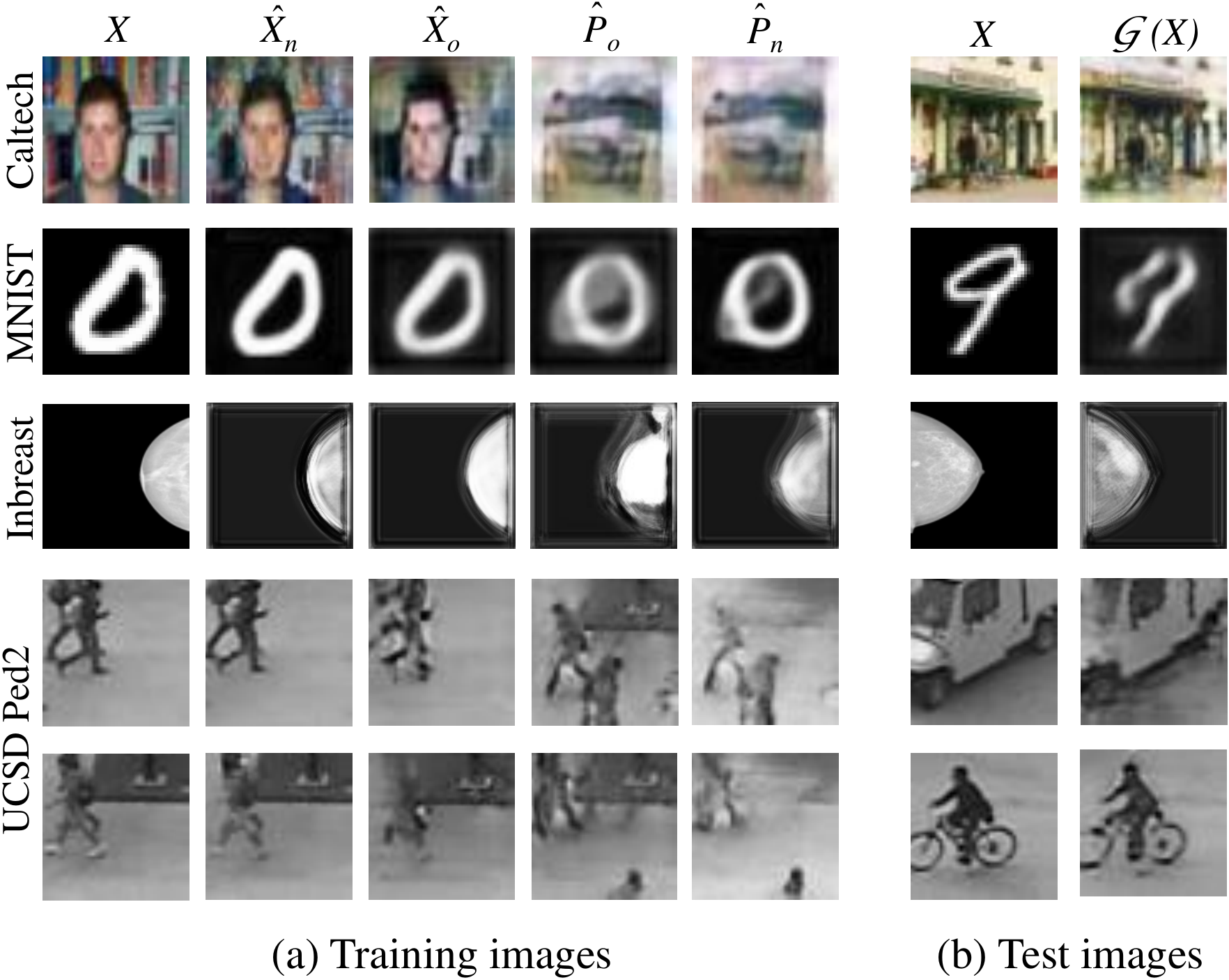}
\end{center}
\caption{Example images taken from different intermediate steps of our framework. (a) Left to right: normal input images ($X$), high quality reconstructions ($\hat{X}_n$), low quality reconstructions ($\hat{X}_{o}$), pseudo anomalies ($\hat{P}_o $), pseudo anomaly reconstructions ($\hat{P}_{n}$). (b) Left column shows outlier/anomaly test images whereas right column shows respective reconstructions $\mathcal{G}_n$($X$).}
\label{fig:example_images}
\end{figure}

\section{Experiments} \label{section:experiments}
\changes{The proposed novelty detection framework for one-class classification is evaluated on six datasets across four different domains including image outlier detection,  video  anomaly  detection,  medical diagnosis and network security. Our approach outperformed several existing state-of-the-art (SOTA) methods across all aforementioned domains which demonstrates its wide scope and significance.}


\noindent\textbf{Evaluation Criteria.} 
We compare our proposed method with current SOTA approaches using area under the curve (AUC) computed at frame level which is a commonly used measure \cite{Nguyen_2019_ICCV,ionescu2019object}. \changes{For some datasets,} we report $F_1$ score and equal error rate (EER) as well \changes{to provide fair comparison with the existing methods} \cite{sabokrou2020deep,ravanbakhsh2019training}.
Additionally, following the SOTA approaches in medical domain \red{\cite{mazhelis2006wbcd, agarap2018breast, dubey2016analysis, oyelade2018st}}, we also report accuracy (ACC) for the medical diagnosis datasets.

\noindent\textbf{Implementation Details.}
We use Adam optimizer \cite{kingma2014adam} with a learning rate of $10^{-3}$ for generator and $10^{-4}$ for discriminator. 
\makesure{$\lambda$ in Eq. \eqref{eq:lambda} is set to $0.2$, $\alpha$  and  $\beta$ in Eq. \eqref{eq:alpha} are set to 0.1 and 0.001.} The value of $\eta$ in Eq. \eqref{eq:eta} is set to 1.25. Moreover, the value of $p$ in Algorithm 1 is set to 3 for KDDCUP and 1 for all other datasets. \revised{For F$_1$ score and EER, $\tau=0.5$ is used in Eq. \eqref{eq:threshold}, while for AUC computation it is varied in the full range of anomaly scores.}




\subsection{Experiments on Images Data}

One of the main applications of  one-class learning algorithms is image outlier detection. Images belonging to the known classes, treated as inliers, are used to train a one-class classification model. Other images that do not belong to these classes are treated as outliers, which the model is supposed to detect based on its training. In this domain, we conducted experiments on two datasets including MNIST and Caltech-256.

\begin{figure}[b]
\begin{center}
   \includegraphics[width=0.75\linewidth]{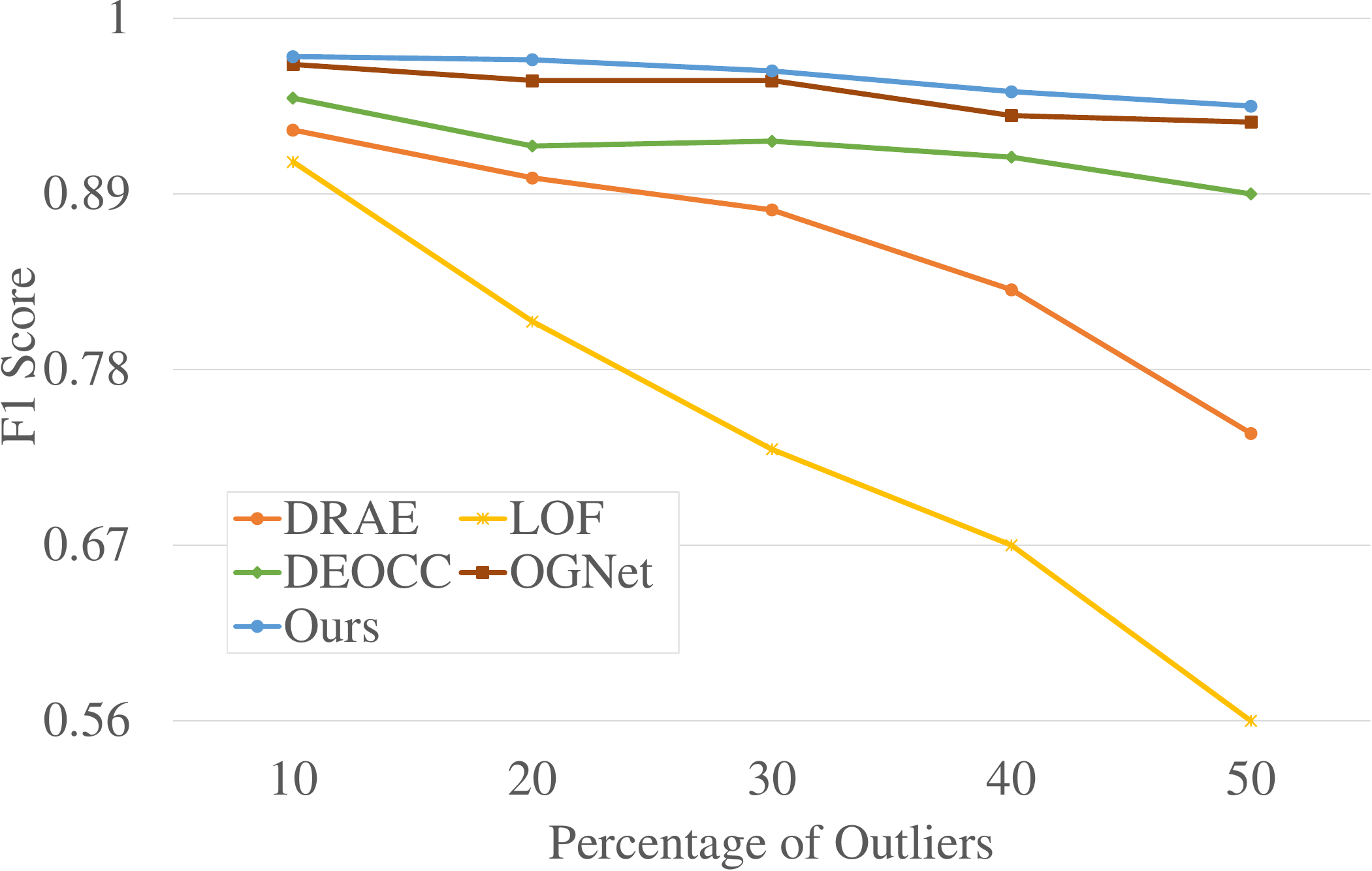}
\end{center}
\caption{\revised{$F_1$ score results on MNIST. Compared to DRAE\cite{xia2015learning_novelty_fig5}, LOF\cite{breunig2000lof_fig5}, OGNet\cite{zaheer2020old}, and DEOCC\cite{sabokrou2020deep}, our approach retains superior performance even with an increased percentage of outliers at test time.}}
\label{fig:f1_score_mnist}
\end{figure}

\subsubsection{MNIST} This dataset \cite{mnist} consists of 60,000 handwritten digits from 0 to 9. Our experimental protocol is kept consistent with the compared SOTA methods \cite{xia2015learning_novelty_fig5,breunig2000lof_fig5,sabokrou2020deep}. In a series of experiments, each category of digits is individually taken as inliers and randomly sampled images of the other categories are considered as outliers. To evaluate the robustness of the compared methods, the ratio of outliers to inliers is increased from 10\% to 50\%.
A comparison of $F_1$ score shows that our approach performs better than the compared methods (Fig.~\ref{fig:f1_score_mnist}). Moreover, the proposed approach demonstrates robust results compared to the existing methods when the percentage of outliers is increased.

\begin{figure}[t]
\begin{center}
   \includegraphics[width=0.75\linewidth]{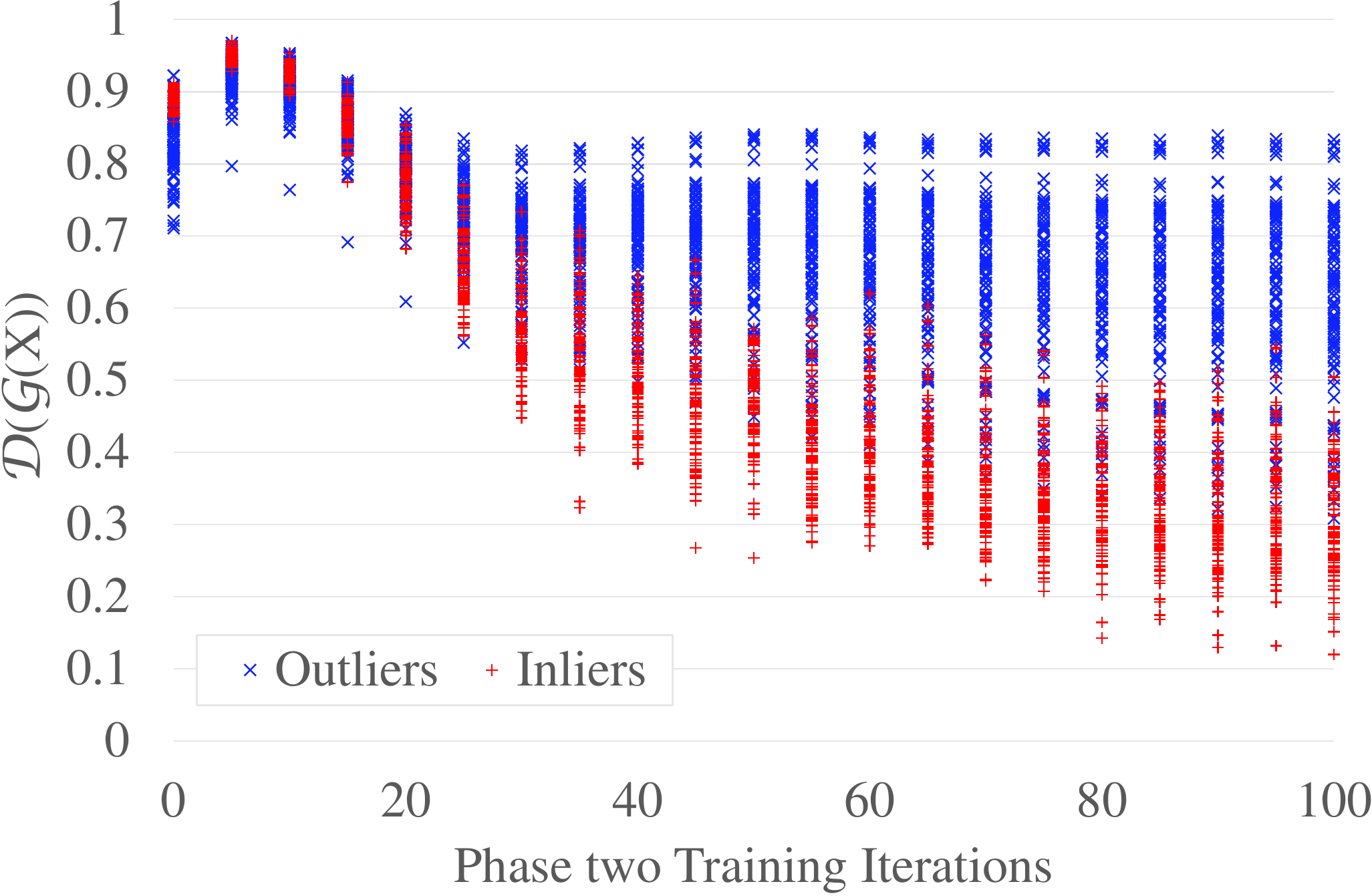}
\end{center}
   \caption{\changes{Variation of anomaly score  distribution over increasing training iterations in phase two of our approach on MNIST dataset. Iteration 0 on x-axis represents the anomaly scores produced by the baseline $\mathcal{G} + \mathcal{D}$ model. Under its conventional role, $\mathcal{D}$ assigns high fake (anomaly) score to the reconstructions of both inliers (red) and outliers (blue) making it difficult to identify each class. Separability of both classes is significantly improved as the phase two training proceeds.} 
}
\label{fig:mnist_score_distribution}
\end{figure}

\begin{figure}[b]
\begin{center}
   \includegraphics[width=0.8\linewidth]{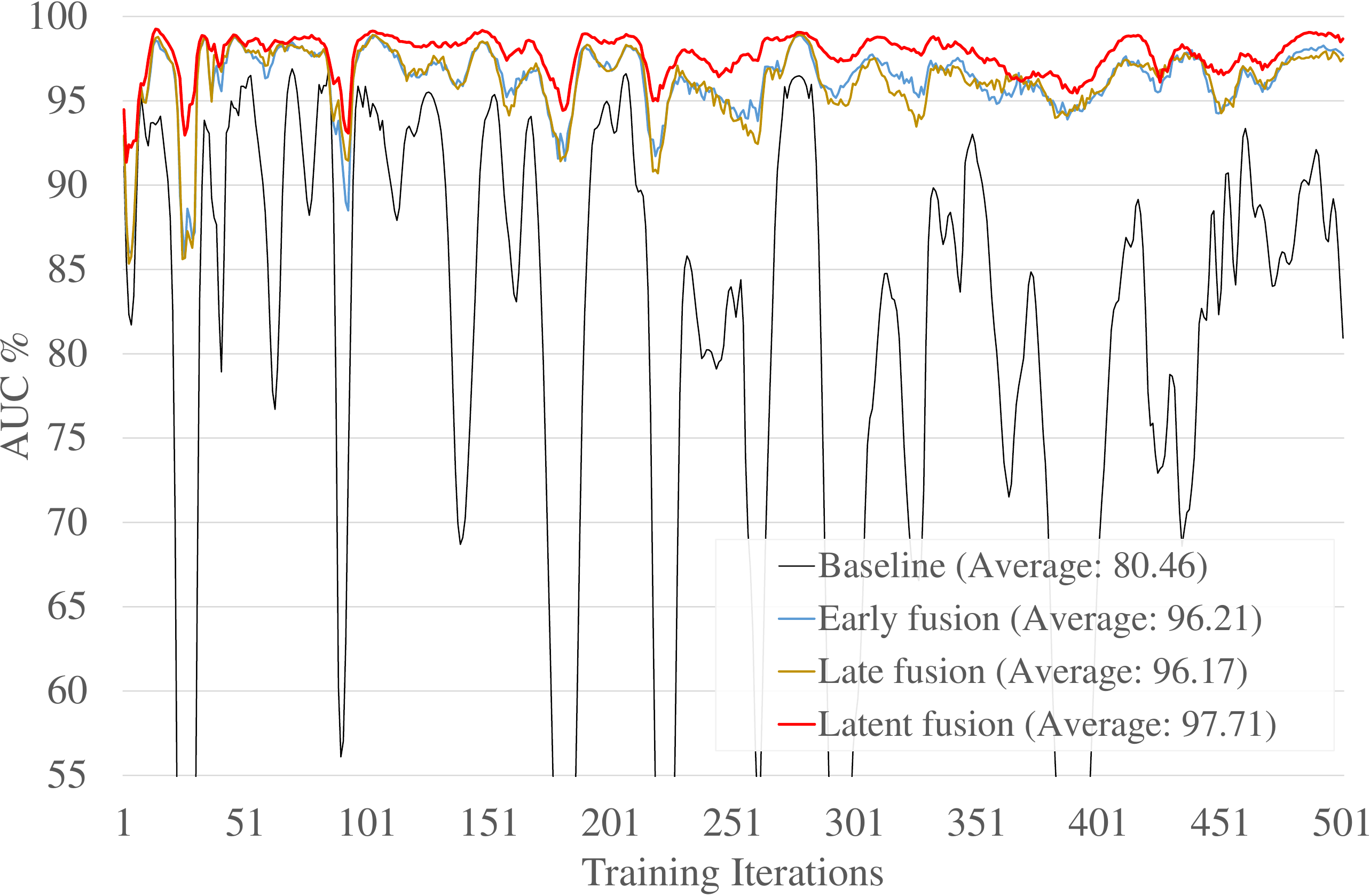}
\end{center}
   \caption{Stability comparison between baseline ($\mathcal{G} + \mathcal{D}$ model) and our approach using three different psuedo anomaly generation modules on MNIST test dataset. In this experiment, $\mathcal{G}_{o}$ is kept fixed, while $\mathcal{G}$ and $\mathcal{D}$ are updated after each iteration of phase one. Then, three separate phase two trainings are followed each with different pseudo anomaly module.}
\label{fig:fusions_comparison_results}
\end{figure}

A step by step analysis of the proposed approach during phase two of the training is shown in Fig.~\ref{fig:mnist_score_distribution}. A significant change in the behavior of the discriminator $\mathcal{D}$ from fake/real (iteration 0 on x-axis) to good/bad quality reconstructions (iterations 1 to 100 on x-axis) can be observed. It may be noted that as the number of iterations increases, the overall distributions of the anomaly scores produced by $\mathcal{D}$ for both classes move apart resulting in more separable outliers. 

Fig. \ref{fig:fusions_comparison_results} shows a performance comparison between the three pseudo anomaly generation modules given by Eq. \eqref{eq:early_fusion}, \eqref{eq:fake_anomaly}, \& \eqref{eq:latent_fusion}. \changes{In order to fairly compare the three approaches, we fixed $\mathcal{G}_{o}$ model after $p = 1$ \changes{and updated $\mathcal{G}$ and $\mathcal{D}$} after each iteration of phase one. Then, followed by each  update, three separate phase two trainings were branched out each with a different pseudo anomaly module.} This way, each configuration is trained on the same baseline models of $\mathcal{G}_n$, $\mathcal{G}_{o}$ and $\mathcal{D}$. It can be observed that our approach with any of the pseudo anomaly generation module is significantly more stable than the baseline $\mathcal{G}$ + $\mathcal{D}$ model. Moreover, on average, the latent fusion based pseudo anomaly module resulted in a better performance and improved stability compared to early fusion and late fusion. 
It is intuitive as reconstructing a corrupt latent representation may result in more natural anomaly examples than fusing images directly.
Therefore, unless stated otherwise, we report all performance results in this manuscript using the latent fusion approach.

\begin{figure}[t]
\begin{center}
   \includegraphics[width=0.8\linewidth]{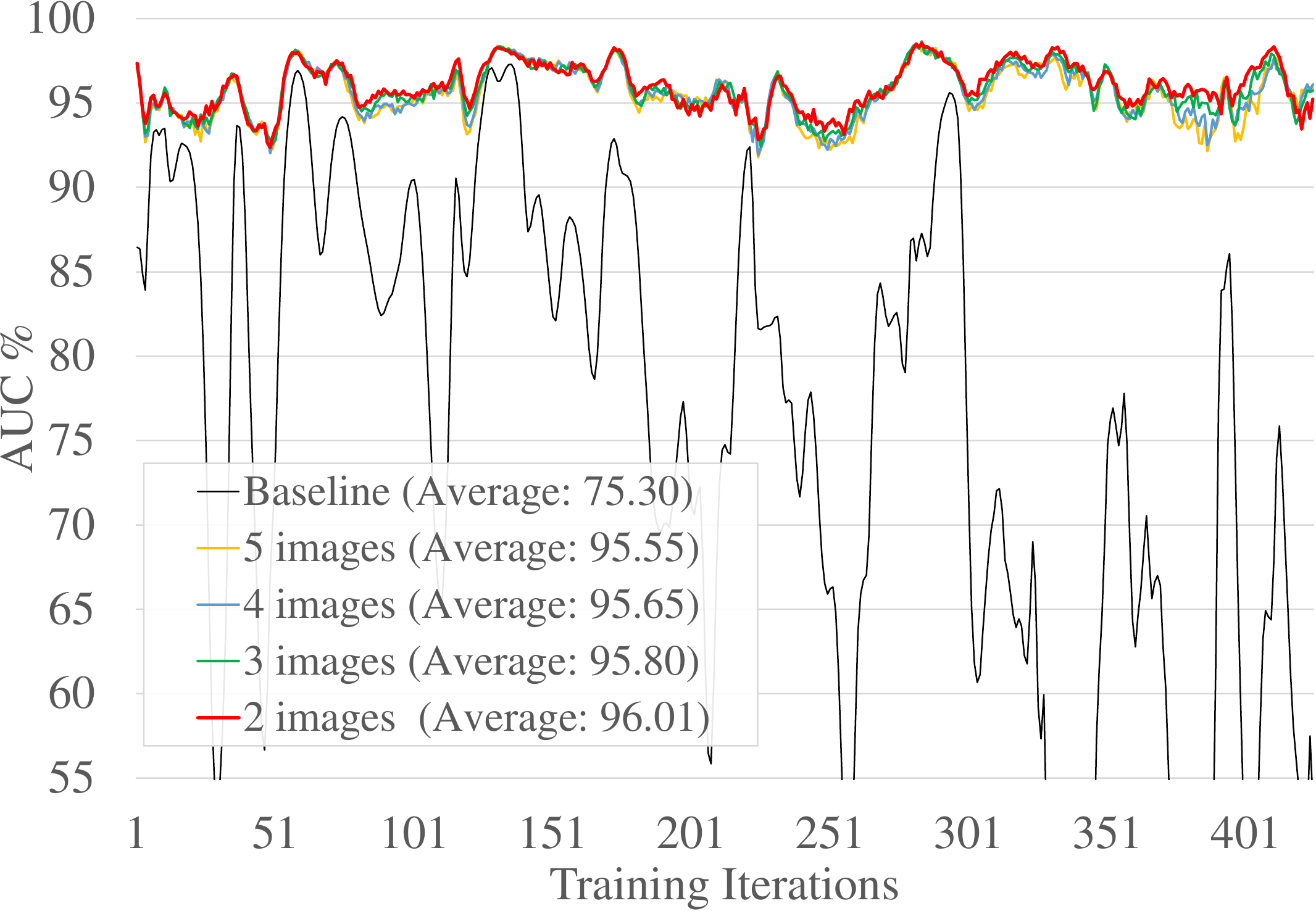}
\end{center}
   \caption{AUC performance comparison of the baseline ($\mathcal{G}+\mathcal{D}$) and our approach using different number of fusion images in the psuedo anomaly module on MNIST test dataset. In this experiment, $\mathcal{G}_{o}$ model is kept fixed and each update in phase one is followed by four separate phase two trainings each with different number of input images in the latent pseudo anomaly module.}
\label{fig:image_mixing_number_comoarison}
\end{figure}

An experiment by varying the number of images to be fused for pseudo anomaly generation is also performed. \changes{The number of images is varied as 2, 3, 4 \& 5 and the AUC performance is plotted in Fig. \ref{fig:image_mixing_number_comoarison}. Similar to the previous experiment, we fixed $\mathcal{G}_{o}$ model after $p = 1$. Then, after each  update } in  phase  one, four separate phase two trainings were branched out, each with different number of images for pseudo anomaly module. It can be seen that the proposed approach with any number of images is significantly more stable than the baseline $\mathcal{G}$ + $\mathcal{D}$ model. Furthermore, it is observed that the performance does not vary significantly when the number of images is changed. However, on average, the two-image based fusion demonstrates slightly better performance. Therefore, all subsequent results reported in this manuscript are computed using the pseudo anomaly module \changes{with two image fusion}.

\begin{table*}[]
\resizebox{\textwidth}{!}{%
\begin{tabular}{c|c|l|ccccccccc}
\multicolumn{1}{l|}{} &
  \multicolumn{2}{l|}{\red{Classes}} &
  DPCP\cite{tsakiris2015dual} &
  REAPER\cite{lerman2015robust} &
  OutlierPersuit\cite{xu2010robust} &
  CoP\cite{rahmani2017coherence} &
  LRR\cite{liu2010robust} &
  R-graph\cite{you2017provable_novelty} &
  DEOCC\cite{sabokrou2020deep} &
  OGNet\cite{zaheer2020old} &
  \green{Ours} \\ \hline
AUC &
  \multicolumn{2}{c|}{\multirow{2}{*}{1}} &
  \begin{tabular}[c]{@{}c@{}}78.3\%\end{tabular} &
  81.6\% &
  83.7\% &
  90.5\% &
  90.7\% &
  94.8\% &
  95.0\% &
  {\ul 98.2\%} &
  \textbf{98.9\%} \\
$F_1$ &
  \multicolumn{2}{c|}{} &
  78.5\% &
  80.8\% &
  82.3\% &
  88.0\% &
  89.3\% &
  91.4\% &
  92.0\% &
  {\ul95.1\%} &
  \textbf{95.5\%}\\ \hline
AUC &
  \multicolumn{2}{c|}{\multirow{2}{*}{3}} &
  79.8\% &
  79.6\% &
  78.8\% &
  67.6\% &
  47.9\% &
  92.9\% &
  93.0\% &
  {\ul97.7\%} &
  \textbf{97.9\%}\\
$F_1$ &
  \multicolumn{2}{c|}{} &
  77.7\% &
  78.4\% &
  77.9\% &
  71.8\% &
  67.1\% &
  88.0\% &
  {\ul93.0\%} &
  91.5\% &
  \textbf{95.2\%}\\ \hline
AUC &
  \multicolumn{2}{c|}{\multirow{2}{*}{5}} &
  67.6\% &
  65.7\% &
  62.9\% &
  48.7\% &
  33.7\% &
  91.3\% &
  93.0\% &
  {\ul98.1\%} &
  \textbf{98.4\%}\\
$F_1$ &
  \multicolumn{2}{c|}{} &
  71.5\% &
  71.6\% &
  71.1\% &
  67.2\% &
  66.7\% &
  85.8\% &
   91.0\% &
  {\ul92.8\%} &
  \textbf{94.7\%}\\ \hline
\end{tabular}%
}
\caption{Performance  comparison of our approach using AUC and $F_1$ score on Caltech-256 \cite{griffin2007caltech} with SOTA methods. Following the existing protocol \cite{you2017provable_novelty}, each subgroup of rows shows evaluation scores on inliers coming from one, three, and five random classes (best and second best performances are shown as bold and underlined respectively).}
\label{tab:auc_f1_caltech}
\end{table*}

\subsubsection{Caltech-256} This dataset \cite{griffin2007caltech} contains a total of 30,607 images belonging to 256 object classes and one `clutter' class. Each category has different number of images, as low as 80 and as high as 827. We followed the same experimental protocol as used by the compared SOTA methods \cite{tsakiris2015dual,lerman2015robust,xu2010robust,rahmani2017coherence,liu2010robust,you2017provable_novelty,sabokrou2020deep}. In a series of three experiments, at most 150 images belonging to \changes{each of the} 1, 3, and 5 randomly chosen classes are defined as training (inlier) data. Outlier images for test are taken from the ‘clutter’ class in such a way that each experiment has exactly equal ratio of outliers and inliers.
The proposed approach depicts an overall superior performance in these experiments in terms of $F_1$ score and AUC, as shown in Table \ref{tab:auc_f1_caltech}. \changes{With the increasing number of inlier classes, our approach demonstrates more consistent robustness than the compared methods.}

\subsection{Experiments on Network Security Dataset}
\changes{To validate the applicability of our proposed approach beyond images, we experimented on a widely popular KDDCUP99 cyber security dataset  from the UCI} repository\footnote{kdd.ics.uci.edu/databases/kddcup99/kddcup99.html}. 
This dataset has been introduced for the purpose of building network intrusion detectors capable of distinguishing between normal connections and intrusions/attacks. It includes a wide variety of intrusions simulated in a military network environment.

Each sample in the original dataset is of 41 dimensions, 34 of which are continuous and 7 are categorical. Following the experimental protocols adopted by Zong \etal \cite{zong2018deep}, we used one-hot representation to encode categorical features and obtained a dataset of 120 dimensions. Since 20\% of the total data samples are “normal” and the rest are labeled as “attack”, “normal” ones are treated as outliers in our experiments.
Half of the inlier class data is used for training and the rest is used for testing along with the outlier class data. The comparison of our approach on KDDCUP dataset provided in Table \ref{Tab:KDDCUP} demonstrates its superiority against SOTA methods \cite{Gong_2019_ICCV,scholkopf2000support,yang2017joint,zong2018deep,zhai2016deep}. Moreover, our approach yields a gain of 5.05\% over the baseline. 
An excellent performance on this dataset demonstrates that our approach is generic and can be effectively applied to the problems in non-visual domains.

\begin{table}[t]
\begin{center}
\begin{tabular}{l|l}
Method & $F_1$ Score \\ \hline
DSEBM \cite{zhai2016deep}  &    0.7399     \\
DCN \cite{yang2017joint}  &   0.7762       \\
OC-SVM \cite{scholkopf2000support} &  0.7954         \\
DAGMM \cite{zong2018deep}  &    0.9369      \\
\revised{OGNet} \cite{zaheer2020old}  &    \revised{0.9590}      \\
MemAE \cite{Gong_2019_ICCV} &    \uline{0.9641}      \\ \hline
Baseline       &    0.9203      \\
Ours     &    \textbf{0.9708}     
\end{tabular}
\caption{Comparison of our approach with current SOTA methods on the network security dataset KDDCUP  (best performance as bold and second best as underlined). }
\label{Tab:KDDCUP}
\end{center}
\end{table}


\begin{table}[b]
\resizebox{\linewidth}{!}{%
\begin{tabular}{l|lcc|l|lcc}
 &
  Method &
  \multicolumn{1}{r}{AUC} &
  \multicolumn{1}{l|}{ACC} &
   &
  Method &
  \multicolumn{1}{l}{AUC} &
  \multicolumn{1}{l}{ACC} \\ \hline
\multirow{12}{*}{\rotatebox{90}{INBreast}} &
  Wei \etal \cite{wei2018inb} &
  84.0 &
  - &
  \multirow{7}{*}{\rotatebox{90}{WBCD$_o$}} &
  Asri \etal \cite{asri2016using} &
  98.3 &
  - \\
 &
  Zhu \etal \cite{zhu2017inb} &
  85.9 &
  \underline{90.0} &
   &
  Support Vector DD\cite{mazhelis2006wbcd}&
  97.3 &
  - \\
 &
  Wu \etal \cite{wu2019inb} &
  89.5 &
  - &
   &
  Naive Parzen\cite{cohen2008wbcd} &
  98.7 &
  - \\
 &
  Ribli \etal \cite{ribli2018inb} &
  \underline{95.0} &
  - &
   &
  Parzen\cite{cohen2008wbcd} &
  99.2 &
  - \\
 &
  Agarwal \etal \cite{agarwal2020inb} &
  90.0 &
  - &
   &
  Typicality Distance\cite{irigoien2008wbcd} &
  \underline{99.4} &
  - \\
 &
   &
   &
   &
   &
  Baseline &
  99.3 &
  \underline{94.1} \\
 &
   &
   &
   &
   &
  Ours &
  \textbf{99.5} &
  \textbf{94.3} \\ \cline{5-8} 
 &
   &
   &
   &
  \multirow{5}{*}{\rotatebox{90}{WBCD$_d$}} &
  Agarap \cite{agarap2018breast} &
  96.1 &
  93.8 \\
 &
   &
   &
   &
   &
  Dubey \etal\cite{dubey2016analysis} &
  - &
  92.0 \\
 &
   &
   &
   &
   &
  Oyelade \etal\cite{oyelade2018st} &
  - &
  81.7 \\
 &
  Baseline &
  85.2 &
  89.5 &
   &
  Baseline &
  \underline{98.8} &
  \underline{95.3} \\
 &
  Ours &
  \textbf{97.5} &
  \textbf{93.0} &
   &
  Ours &
  \textbf{98.9} &
  \textbf{95.6}
\end{tabular}%
}
\caption{AUC and ACC performance comparisons on medical diagnosis datasets: Inbreast, Wisconsin Breast Cancer Original (WBCD$_o$) and Wisconsin Breast Cancer Diagnostic (WBCD$_d$). Best performances in each dataset are mentioned in bold whereas second best are underlined.}
\label{tab:INB-table}
\end{table}

\subsection{Experiments on Medical Diagnosis Datasets}
In medical diagnosis, we experimented on two datasets including Inbreast \cite{moreira2012inbreast} and Breast Cancer Wisconsin \cite{Wolberg1992wbcd}. 
\subsubsection{Inbreast} This dataset consists of 410 full-field digital mammography images obtained from 115 subjects~\cite{moreira2012inbreast}. These images consist of several types of lesions such as masses, calcifications, asymmetries, and distortions.  Following the existing protocol \cite{zhu2017inb}, we  used 67 images of class BI-RADS 1 (benign) and 49 images of class BI-RADS 5 (malignant)  for our experiments. For training 59 benign images were used while for testing, the remaining 8 benign and all of the malignant images were utilized. 

The proposed approach is compared with the existing SOTA methods \cite{wei2018inb,zhu2017inb,wu2019inb,ribli2018inb,agarwal2020inb} and our baseline ($\mathcal{G} + \mathcal{D}$) model. The results summarized in Table \ref{tab:INB-table} show that our approach outperforms the compared recent methods by a significant margin.  

\subsubsection{Breast Cancer Wisconsin Dataset} 
This dataset has two different versions, including Wisconsin Breast Cancer-Original\footnote{archive.ics.uci.edu/ml/datasets/breast+cancer+wisconsin+(original)} (WBCD$_o$) and Wisconsin Breast Cancer-Diagnostic\footnote{archive.ics.uci.edu/ml/datasets/Breast+Cancer+Wisconsin+(Diagnostic)}  (WBCD$_d$). \noindent{WBCD$_o$} consists of 683 instances each having ten attributes describing characteristics of the cell nuclei present in the image~\cite{Wolberg1992wbcd}. Each instance belongs to either malignant (239 samples) or normal  class (444 samples).  Half of the normal instances are used for the training and the rest of the normal and all malignant samples are used for testing purpose. 
\noindent{WBCD$_d$} consists of 569 instances each having 30 real valued attributes. Each instance belongs to either malignant (212 samples) or normal class (357 samples). 70\% of the normal instances are used for training and the rest of the normal and all malignant samples are used for testing.
Following the existing SOTA approaches, AUC and accuracy  comparisons are provided for both versions of the dataset. 
\changes{The results summarized in Table \ref{tab:INB-table} show that our approach demonstrates superior performance on both categories of the WBCD dataset.}


\subsection{Anomaly Detection in Videos}
One-class classifiers have often been used in the domain of anomaly detection for surveillance purposes \cite{sultani2018real_novelty,tudor2017unmasking_novelty,dutta2015online,zhang2016video_novelty,ravanbakhsh2019training}.
This task is more complicated than outlier detection because of the involvement of moving objects which may cause variations in appearance.
In this domain we evaluated our method on UCSD Ped2 \cite{chan2008ucsd} as discussed below.

\begin{table}[]
\resizebox{\linewidth}{!}{%
\begin{tabular}{cccc}
LDRAM\cite{xu2015learning_denoise} &RE\cite{sabokrou2016video} & TAD\cite{ravanbakhsh2019training} & AbGAN\cite{ravanbakhsh2017abnormal_novelty}       \\ \hline
17\%     &  15\%    & 14\%   &  13\%    \\ \hline\hline
  DEOCC\cite{sabokrou2020deep}   & DC\cite{sabokrou2017deep_novelty}   & OGNet \cite{zaheer2020old}   & Ours         \\ \hline
12.5\% & {\ul 9\%}         & \textbf{7\%}  & \textbf{7\%}
\end{tabular}}
\caption{EER results comparison with existing works on UCSD Ped2 dataset. Lower numbers mean better results. 
Best performance is highlighted as bold while second best as underlined.
}
\label{Tab:EER}
\end{table}

\subsubsection{USCD Ped2} This dataset  comprises of 2,550 frames in 16 training videos (normal frames) and 2,010 frames in 12 test videos (containing both normal and abnormal frames) \cite{chan2008ucsd}. Pedestrians dominate most of the frames whereas anomalies include skateboards, vehicles, bicycles, etc.
\changes{Spatial size of the anomaly objects appearing in this dataset is relatively smaller than the overall frame size therefore, following Sabokrou \etal \cite{sabokrou2020deep}, we extract several patches from the video frames.} \blue{Specifically, we divide each \changes{frame $I$ of size $240 \times 360$ of the Ped2 dataset} into grayscale non-overlapping patches $X_{I} = \{X_1, X_2, ... , X_n\}$ of resolution $45 \times 45$ pixels. Normal videos, containing only the scenes of walking pedestrians, are used to extract patches for training.} Test patches are extracted from abnormal videos which contain abnormal as well as normal scenes. 
\blue{Anomaly score of a frame is determined by the maximum anomaly score of its corresponding patches as:} 
\begin{equation}\label{eq:frame_anomaly_score}
A_{I} = \underset{X}{\text{max}}\:\mathcal{D}(\mathcal{G}(X)) \text{, where $X \in X_{I}$}
\end{equation}
In order to minimize the unnecessary inference of test patches, a frame difference based motion detection criterion is set to exclude patches without motion i.e. patches containing background pixels. \changes{Specifically, if pixel values of the current patch do not change at least 20 \% from its corresponding patch in a neighboring frame, we discard the patch.}
Using frame-level AUC and EER as the two evaluation metrics, we compare our approach with a series of existing works \cite{tudor2017unmasking_novelty,luo2017revisit_novelty,nguyen2019hybrid,hinami2017joint_novelty,liu2018classifier_novelty,luo2017remembering,Gong_2019_ICCV,ravanbakhsh2018plug_novelty,sun2017online,hasan2016learning_novelty,liu2018future_novelty,xu2015learning_denoise,Nguyen_2019_ICCV,zhang2016video_novelty,ionescu2019object,zhao2017spatio,sabokrou2020deep} published within last 6 years. The corresponding results provided in Tables \ref{Tab:EER} and \ref{tab:ped2AUC} demonstrate the superiority of our approach against all compared methods. Moreover, our approach yields a gain of 5.2\% in AUC when compared with the baseline ($\mathcal{G} + \mathcal{D}$ model). Few examples of the reconstructions are provided in Fig.~\ref{fig:example_images}. As shown in Fig.~\ref{fig:example_images}b, good reconstructions are often generated by $\mathcal{G}$ however, our framework still successfully identifies anomalous inputs. It is majorly because of our proposed pseudo anomaly module which enables $\mathcal{D}$ to learn the underlying behavior of $\mathcal{G}$ on anomalous images. It also explains the consistent performance of our proposed approach across a wide range of training iterations compared to the baseline \changes{ (Fig.~\ref{fig:AUC_Comparison}, \ref{fig:fusions_comparison_results} \& \ref{fig:image_mixing_number_comoarison}).}

\begin{table}[]
\resizebox{\linewidth}{!}{%
\begin{tabular}{ll|ll}
Method                                    & \multicolumn{1}{c|}{AUC} & Method                             & \multicolumn{1}{c}{AUC} \\ \hline
Unmasking\cite{tudor2017unmasking_novelty}& 82.2\%                   & MemAE\cite{Gong_2019_ICCV}         & 94.1\%             \\
HybridDN\cite{nguyen2019hybrid}           & 84.3\%                   & GrowingGas\cite{sun2017online}     & 94.1\%                  \\
Liu \etal\cite{liu2018classifier_novelty} & 87.5\%                   & FFP\cite{liu2018future_novelty}    & 95.4\%                  \\
ConvLSTM\-AE\cite{luo2017remembering}     & 88.1\%                   & Abati \etal\cite{abati2019latent}  & 95.8\%                  \\
Ravanbakhsh \etal\cite{ravanbakhsh2018plug_novelty} & 88.4\%         & ConvAE+UNet\cite{Nguyen_2019_ICCV} & 96.2\%             \\
ConvAE\cite{hasan2016learning_novelty}    & 90\%                     & STAN\cite{lee2018stan}             & 96.5\%                  \\
AMDN\cite{xu2015learning_denoise}         & 90.8\%                   & Chang \etal\cite{chang2020clustering}& 96.5\%                  \\
Hashing Filters\cite{zhang2016video_novelty}        & 91\%           & BMAN\cite{lee2019bman}             & 96.6\%             \\
AE\_Conv3D\cite{zhao2017spatio}           & 91.2\%                   & MGNAD\cite{park2020learning}       & 97.0\%                   \\
TSC\cite{luo2017revisit_novelty}          & 92.2\%                   & RAD\cite{vu2019robust }                   & 97.5\% \\
FRCN action\cite{hinami2017joint_novelty} & 92.2\%                  & Object-centric\cite{ionescu2019object}            & 97.8\%  \\
AbnormalGAN\cite{ravanbakhsh2017abnormal_novelty} & 93.5\%                 & OGNet\cite{zaheer2020old} &    \underline{98.1\%}     \\ \hline
Baseline                                  & 92.9\%                   &  \green{Ours}                               & \textbf{98.3\%}        
\end{tabular}%
}
\caption{\blue{AUC performance comparison of our approach on UCSD Ped2 dataset with SOTA works published in the last 6 years. Best performance is bold and second best is underlined.}}
\label{tab:ped2AUC}
\end{table}

\subsection{Analysis}
\label{sec:discussion}

\begin{table}[b]
\begin{center}
\begin{tabular}{cccc}
  &  \green{Minimum} & \green{Maximum} & \green{Average $\pm$ STD}         \\ \hline
OGNet \cite{zaheer2020old}    & 93.13    &  97.97    & 96.37 $\pm$ 1.40  \\
Ours                          & \textbf{95.49}   & \textbf{98.79} & \textbf{97.91 $\pm$ 1.06}
\end{tabular}
\caption{Minimum, maximum and average $\pm$ standard deviation AUC \% performance comparison of OGNet and our current approach on a series of ten experiments using MNIST dataset. }
\label{Tab:ComparisonOursOGNET}
\end{center}
\end{table}

\green{\noindent\textbf{Comparison with OGNet:} 
In addition to the comparisons between our current approach and OGNet\cite{zaheer2020old} provided in Fig. \ref{fig:fusions_comparison_results} and Tables \ref{tab:auc_f1_caltech}, \ref{Tab:EER}, \& \ref{tab:ped2AUC}, we report in-depth performance and stability analysis of both approaches. In a series of ten experiments using MNIST, an average, minimum and maximum of the AUC performance of both OGNet and our current approach is reported in Table \ref{Tab:ComparisonOursOGNET}. As seen, our approach performs exceptionally well, demonstrating superior stability across several experiments. Overall, the improvement in performance can be attributed to the usage of latent fusion (Eq. \eqref{eq:latent_fusion}) for pseudo anomalies generation and the automated selection of $\mathcal{G}_o$ and $\mathcal{G}_n$ (Algorithm 1).}

\revised{In addition to the above analysis, we also evaluate the effectiveness of the usage of $\eta$ by incorporating it into OGNet \cite{zaheer2020old}. It resulted in an average AUC of $97.06\%$, which is better than that of OGNet ($96.37\%$) but lower than our proposed approach ($97.91\%$), verifying the importance of the proposed stopping criteria.}


\begin{figure}[t]
\begin{center}
  \includegraphics[width=0.8\linewidth]{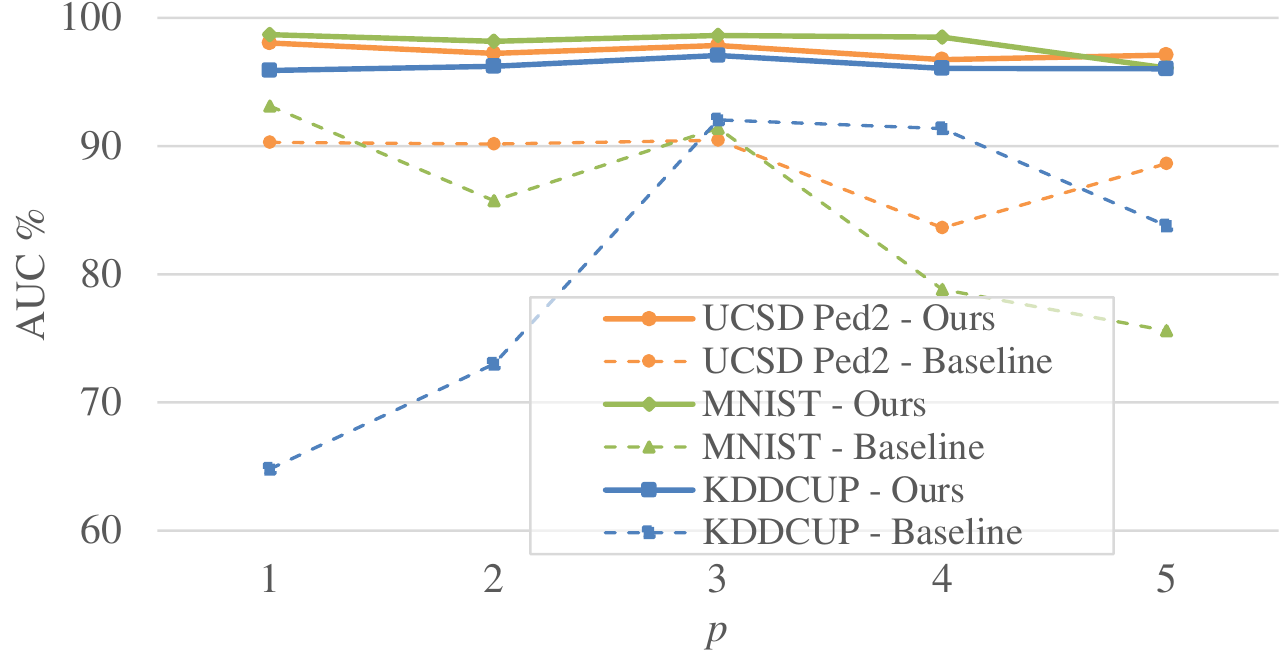}
\end{center}
  \caption{ \revised{The plot shows AUC performance comparison of our approach (solid lines) with the baseline (dotted lines) on three different datasets} with varying \red{$p$} in Algorithm 1. $p$ defines the number of burn-in epochs in phase one before we start the selection of $\mathcal{G}_{o}$ and $\mathcal{G}_{n}$. Optimal performance is achieved when $p$ is set to 1 however, higher values of $p$ also yield comparable results. Overall, our approach consistently demonstrates higher as well as more stable results compared to the baseline.}
\label{fig:varying_n}
\end{figure}

\begin{figure}[b!]
\begin{center}
  \includegraphics[width=0.8\linewidth]{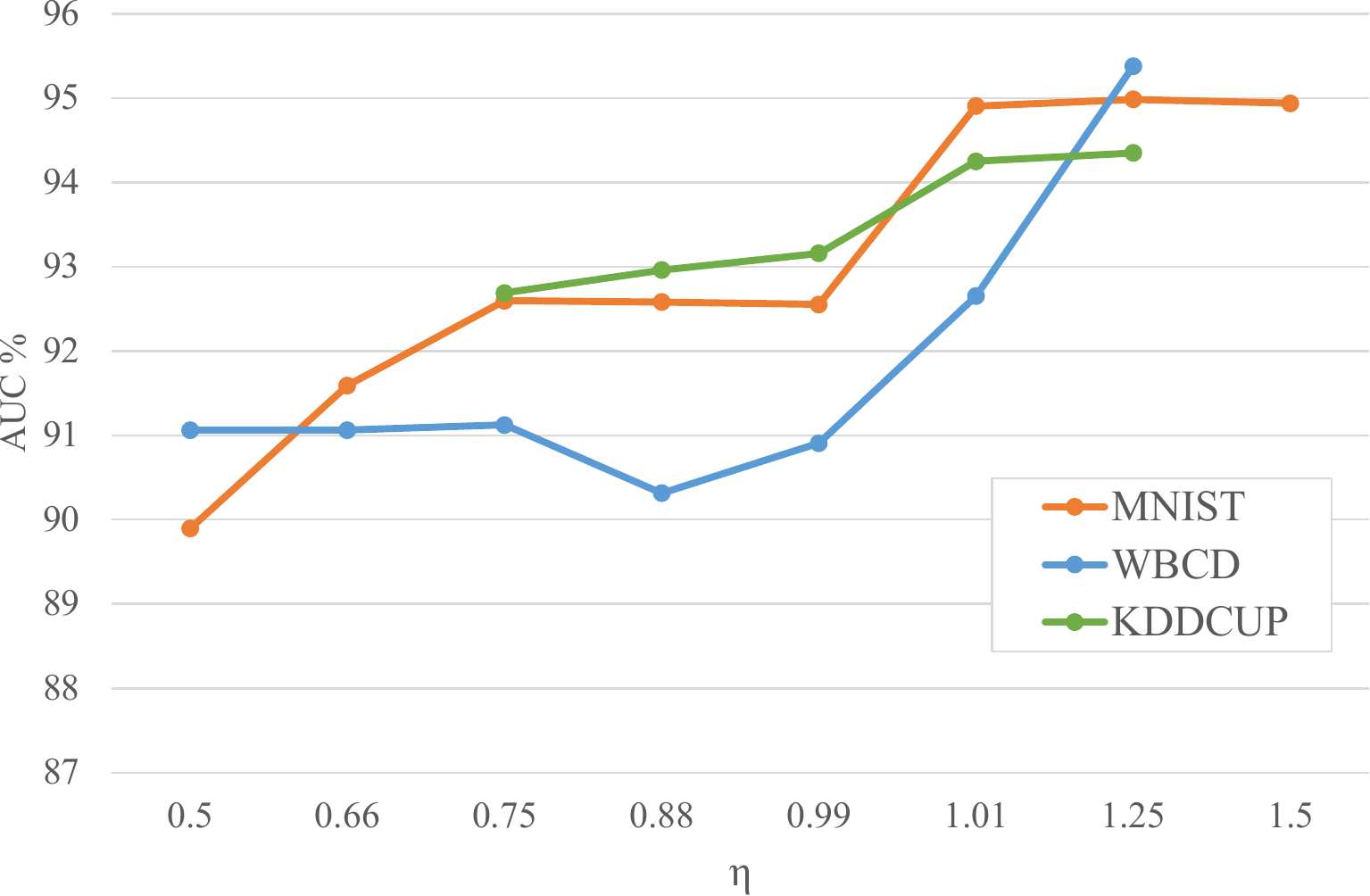}
\end{center}
  \caption{\green{The threshold $\eta$ on the ratio of the reconstruction losses between $\mathcal{G}_{o}$ and $\mathcal{G}_{n}$ (Eq. \eqref{eq:eta}) is plotted on the horizontal axis against the average AUC \% performance on the vertical axis for MNIST, WBCD and KDDCUP datasets. It can be seen that with the increase in values of $\eta$, the overall performance of our system demonstrates an increasing trend. For some datasets however, due to the maximum number of training epochs or the limited reconstruction capability of the generator network, a certain $\eta$ value may never reach. In such cases, we do not report any number. Nevertheless, $\eta = 1.25$ was always achievable on all datasets.}}
   

\label{fig:recon_analysis}
\end{figure}

\noindent\textbf{When to start the selection of $\mathcal{G}_{o}$ and $\mathcal{G}_{n}$?} 
\changes{The performance of our framework is not strictly dependent on the selection of $p$ i.e., the number of epochs in the burn-in stage before we begin to make a selection of $\mathcal{G}_{o}$ and $\mathcal{G}_{n}$ in Algorithm 1.} Fig.~\ref{fig:varying_n} shows the AUC performance of phase two training at varying values of $p$ on MNIST, Ped2 and KDDCUP datasets. \changes{This selection of datasets is arbitrary with an intent to limit the number of computationally expensive experiments.} For Ped2 and MNIST, $p = 1$ demonstrates slightly superior results compared to other values of $p$, whereas for KDDCUP, $p = 3$ yields slightly better performance. Overall, the performance is almost identical for all values of $p$ on the under evaluation datasets except on MNIST where the system demonstrates slightly deteriorated results at $p = 5$. This may be attributed to a slight over-training of the $\mathcal{G}_{o}$ network, as MNIST is a relatively smaller dataset. Intuitively, lower $p$ value is desirable as it may allow our framework to obtain higher $\eta$ values, thus improving the overall performance.



\noindent\textbf{When to stop phase one training?} 
Once $\mathcal{G}_{o}$ is selected, phase one training continues until the reconstruction loss ratio $\eta$, as described in Eq. \eqref{eq:eta}, is obtained. At this point, we obtain $\mathcal{G}_{n}$ and begin phase two. We empirically observed that for values of $1 \le \eta \le 1.25 $ the phase one training terminates within few epochs.

\noindent\textbf{Which value of $\eta$ to be selected?}
As shown in Fig. \ref{fig:recon_analysis}, our approach performs significantly better for values of $\eta > 1.00$. \changes{As described in Eq. \eqref{eq:eta}, $\eta$ is the threshold on the ratio of the reconstruction losses of $\mathcal{G}_{o}$ and $\mathcal{G}_{n}$. Therefore, $\eta > 1$ means $\mathcal{G}_{n}$ is ensured to have a better reconstruction quality compared to  $\mathcal{G}_{o}$.} Moreover, increasing the values of $\eta$ further beyond 1.00 demonstrates an increasing trend in performance. 
Values of $\eta$ significantly less than one, which means the reconstruction loss of $\mathcal{G}_{o}$ is lower than $\mathcal{G}_{n}$, may result in a degraded system performance. Although, our proposed approach tries to handle it well due to the presence of pseudo anomaly module and phase two training, difference in the overall performance is noticeable. An interesting phenomenon can be observed  when the $\eta$ value is changed from 0.99 to 1.01 where our approach demonstrates a swift elevation in the performance for all under observation datasets. This further validates our hypothesis that  $\eta > 1$ is likely to improve the overall system performance. However, it may be noted that a value of $\eta$ significantly greater than $1.00$ may never be achieved due to several factors such as limited reconstruction capability of the generator model, smaller dataset, higher $n$ values etc. We empirically found that $\eta = 1.25$ was achievable in all experiments across each of the four domains.

\begin{table}[t]
\resizebox{\linewidth}{!}{
\begin{tabular}{c|c|cccccc}   & Phase one  & \multicolumn{5}{c}{Phase two }               \\ \hline
$X$                          & \ding{51}  & -               & \ding{51}     & \ding{51}     & \ding{51}     & \ding{51}     & \ding{51}              \\
$\hat{X}_n$                  & \ding{51}  & \ding{51}       & \ding{51}     & \ding{51}     & \ding{51}     & \ding{51}     & \ding{51}              \\
$\hat{X}_{o}$                & -          & \ding{51}       & \ding{51}     & -             & \ding{51}     & -             & \ding{51}              \\
$\hat{P}_{n}$                & -          & -               & -             & \ding{51}     & -             & \ding{51}     & \ding{51}              \\
\revised{$\hat{P}_o$}                  & -          & -               & -             & -             & \ding{51}             & \ding{51}     & -                \\ \hline
AUC                          & 92.9\%     & 94.4\%          & 95.1\%        & 96.6\%        & 88.5\%        & \revised{90.2\%}        & \textbf{98.3\%}
\end{tabular}
}
\caption{Ablation of our framework on UCSD Ped2 dataset.} 
\label{tab:ablation}
\end{table}

\subsection{Ablation}
Ablation study of our approach on Ped2 is provided in Table \ref{tab:ablation}. As seen, all components of our proposed model including real images $X$, low quality reconstructions $\hat{X}_{o}$, high quality reconstructions $\hat{X}_{n}$, and pseudo anomaly reconstructions $\hat{P}_{n}$ contribute significantly towards a robust training. Moreover, it may be noted that with either of these components missing, our network still yields superior AUC than the baseline. \revised{Some interesting insights can be observed in the second and the third columns from the right, where pseudo anomaly $\hat{P}_{o}$ is directly input to $\mathcal{D}$ as part of the bad quality reconstruction examples. Two possible cases are explored: 1) The performance is computed after excluding the  pseudo anomaly reconstruction step and providing $\hat{P}_{o}$ directly to $\mathcal{D}$. 
2) We exclude $\hat{X}_{o}$ and instead, use $\hat{P}_{n}$ and $\hat{P}_{o}$ together as bad quality examples.}
\revised{As seen, both these configurations result in a significant drop in performance.
These configurations perform even worse than the baseline, showing $88.5\%$ and $90.2\%$ AUC compared to $92.9\%$ of the baseline. The performance drop can be well-attributed to the involvement of $\hat{P}_{o}$ to train $\mathcal{D}$ directly.} This shows the importance of our overall pseudo anomaly generation pipeline. Once fake anomalies are created, it is necessary to reconstruct them using $\mathcal{G}_{n}$. Intuitively, it is understandable, as during testing only $\mathcal{G}_{n}$ and $\mathcal{D}$ are utilized for anomaly scoring. The reconstruction of pseudo anomalies helps $\mathcal{D}$ to learn the  underlying behavior of  $\mathcal{G}_{n}$ in reconstructing anomalous data, thus resulting in a more robust one-class classification system.

\section{Conclusion}
\blue{The paper presented a novel formulation that utilizes both the generator ($\mathcal{G}$) and the discriminator ($\mathcal{D}$) of an adversarial network to perform anomaly detection. Conventional $\mathcal{G}$ and $\mathcal{D}$ unified models used towards such problems often demonstrate unstable performance due to adversarial training. We attempted to transform the ultimate role of $\mathcal{D}$ from identifying real and fake to classifying good and bad quality reconstructions, an approach that aligns well with the anomaly detection using generative networks. A pseudo anomaly module is also proposed to create fake anomaly examples from the normal training data. These fake anomalies help $\mathcal{D}$ in learning about the behavior of $\mathcal{G}$ on unusual input data. Furthermore, we also devise a generic stopping criterion that ensures optimal convergence. Our extensive experimentation on six challenging datasets demonstrate the superiority of our approach across four different domains i.e. image outliers detection, video anomaly detection, medical diagnosis, and cybersecurity.}

\ifCLASSOPTIONcompsoc
  \section*{Acknowledgments}
\else
  \section*{Acknowledgment}
\fi

 This work was supported by the ICT R\&D program of MSIP/IITP. [2017-0-00306, Development of Multimodal Sensor-based Intelligent Systems for Outdoor Surveillance Robots].

\ifCLASSOPTIONcaptionsoff
  \newpage
\fi



%


\bibliographystyle{IEEEtran} 
\bibliography{main_v_2}


%

\vspace{-15mm}
\begin{IEEEbiography}[{\includegraphics[width=1in,height=1.25in,clip,keepaspectratio]{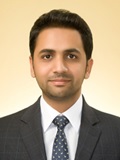}}]{Muhammad Zaigham Zaheer}
is currently associated with the Electronics and Telecommunications Research Institute (ETRI), Ulsan, Korea, as a post-doc researcher. The research work presented in this paper was carried out during his PhD at the University of Science and Technology, Daejeon, Korea. At that time he was also affiliated with ETRI as a student researcher. Previously, he received his MS degree from Chonnam National University, Gwangju, Korea, in 2017 and his undergraduate degree from Pakistan Institute of Engineering and Applied Sciences, Islamabad, Pakistan, in 2012. His current research interests include computer vision,  anomaly detection in images/videos, semi-supervised/self-supervised learning, and video object segmentation.
\end{IEEEbiography}
\vspace{-15mm}
\begin{IEEEbiography}[{\includegraphics[width=1in,height=1.25in,clip,keepaspectratio]{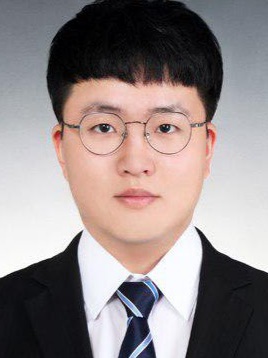}}]{Jin Ha Lee}
Jin-Ha Lee received his BS in computer engineering from Inha University, Incheon, Korea, in 2019. He is currently working towards his MS degree in computer software at the University of Science and Technology (UST), Daejeon, Korea. His recent interests include deep learning, computer vision, and anomaly detection.
\end{IEEEbiography}
\vspace{-15mm}
\begin{IEEEbiography}[{\includegraphics[width=1in,height=1.25in,clip,keepaspectratio]{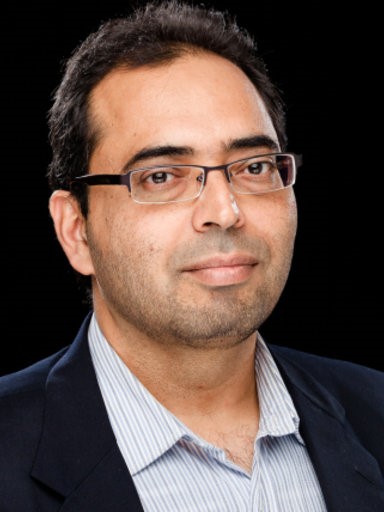}}]{Arif Mahmood}
Arif Mahmood is a Professor in the Department of Computer Science, Information Technology University, Lahore, Pakistan. He also worked as Research Assistant Professor in the University of Western Australia and PostDoc Researcher in Qatar University. His broad areas of interest include Computer Vision and Machine Learning applications. He has  worked on moving objects detection in videos, visual object tracking, visual object categorization, nucleus detection and tissue phenotyping in cancer histology images, face detection and facial expression synthesis, action detection and recognition, visual crowd analysis, anomalous event detection,  human body pose estimation, sparse spectral clustering, community detection in complex networks, ocean color monitoring using remote sensing, and applications of Machine Learning in Cloud Computing.
\end{IEEEbiography}
\vspace{-15mm}
\begin{IEEEbiography}[{\includegraphics[width=1in,height=1.25in,clip,keepaspectratio]{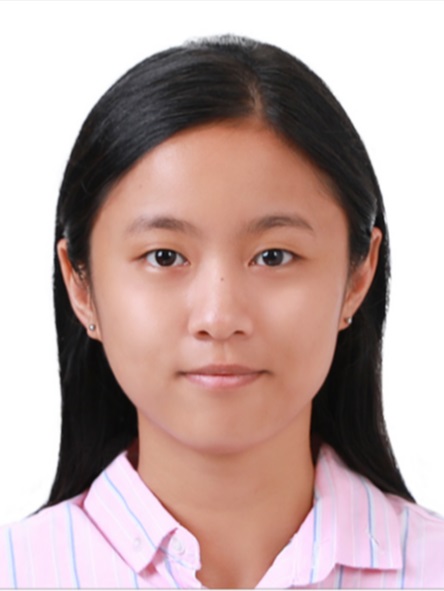}}]{Marcella Astrid}
Marcella Astrid received her BEng in computer engineering from the Multimedia Nusantara University, Tangerang, Indonesia, in 2015, and the MEng in computer software from the University of Science and Technology (UST), Daejeon, Korea, in 2017. At the same university, she is currently working towards her PhD degree in computer science. Her recent interests include deep learning and computer vision.
\end{IEEEbiography}
\vspace{-15mm}
\begin{IEEEbiography}[{\includegraphics[width=1in,height=1.25in,clip,keepaspectratio]{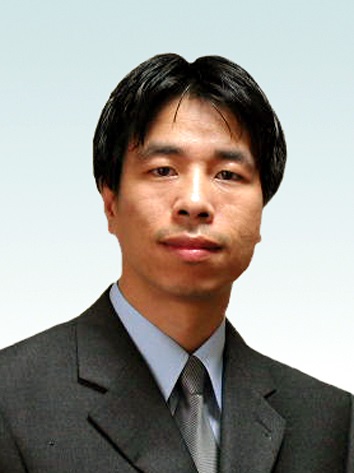}}]{Seung-Ik Lee}
Seung-Ik Lee received his BS, MS, and PhD degrees in computer science from Yonsei University, Seoul, Korea, in 1995, 1997 and 2001, respectively. He is currently working for the Electronics and Telecommunications Research Institute, Daejeon, Korea. Since 2005, he has been with the Department of Computer Software, University of Science and Technology, Daejeon, Korea, where he is now a professor. His research interests include  computer vision, deep learning, and reinforcement learning.
\end{IEEEbiography}






\end{document}